\documentclass[twoside]{article}

\usepackage[accepted]{aistats2022}
\input{my_commands.sty}

\usepackage[round]{natbib}

\newcommand{\affil}[1]{\textsuperscript{\textnormal{#1}}} 

\begin{document}

\runningtitle{Discovering Inductive Bias with Gibbs Priors}

\twocolumn[

\aistatstitle{
Discovering Inductive Bias with Gibbs Priors: \\
A Diagnostic Tool for Approximate Bayesian Inference
}

\aistatsauthor{Luca Rendsburg\affil{1} \And Agustinus Kristiadi\affil{1} \And  Philipp Hennig\affil{1,2} \And Ulrike von Luxburg\affil{1,2}}
\runningauthor{Luca Rendsburg, Agustinus Kristiadi,  Philipp Hennig, Ulrike von Luxburg}

\aistatsaddress{%
\affil{1}University of T\"ubingen \hspace{50pt}
\And
\affil{2}Max Planck Institute for Intelligent Systems, T\"ubingen
} 
]

\begin{abstract}
Full Bayesian posteriors are rarely analytically tractable, which is why real-world Bayesian inference heavily relies on approximate techniques.
Approximations generally differ from the true posterior and require diagnostic tools to assess whether the inference can still be trusted.
We investigate a new approach to diagnosing approximate inference: the approximation mismatch is attributed to a change in the inductive bias by treating the approximations as exact and reverse-engineering the corresponding prior.
We show that the problem is more complicated than it appears to be at first glance, because the solution generally depends on the observation.
By reframing the problem in terms of incompatible conditional distributions we arrive at a natural solution: the \emph{Gibbs prior}.
The resulting diagnostic is based on pseudo-Gibbs sampling, which is widely applicable and easy to implement.
We illustrate how the Gibbs prior can be used to discover the inductive bias in a controlled Gaussian setting and for a variety of Bayesian models and approximations.
\end{abstract}

\section{INTRODUCTION}\label{sec:introduction}
Bayesian inference is based on the posterior distribution $p(\latent|\obs)$ over latent variables $\latent$ given an observation $\obs$.
Bayes' theorem gives an explicit formula for computing the posterior, but is often infeasible in practice because the latent space is too large to work with, the appearing integrals are intractable, or the likelihood function cannot be evaluated.
In these cases, practitioners revert to approximating the posterior instead.
This approach comprises a cornucopia of methods, which can be divided into two groups.
The first group consists of deterministic approximation methods that compute a feasible approximating distribution
\footnote{While standard notation for the approximation is $\VIapprox(\latent)$, it will be useful in the context of this paper to think of it as a conditional distribution.}
$\VIapprox(\latent|\obs)$ to the exact posterior $\posterior(\latent|\obs)$ and includes methods such as variational inference \citep{Hin:1993, Jor:1999, Ble:2017, Hof:2013, Ran:2014, Kuc:2017}, Laplace approximations \citep{Spi:1990, Mac:1992, Rue:2009, Rue:2017, Dax:2021}, and expectation propagation \citep{Min:2001}.
The second group consists of stochastic sampling methods that generate samples from (an approximation to) the posterior and includes methods such as Markov chain Monte Carlo \citep{Cas:1992, Hof:2014, Bar:2017} and approximate Bayesian computation \citep{Dig:1984, Sis:2018, Bea:2019}.
For a general introduction to approximate methods in Bayesian inference see \citet{Bis:2006}.
While approximate methods make Bayesian inference feasible, they come at the cost of a distortion in the posterior.
The resulting approximate inference can deviate significantly from exact Bayesian inference.
This calls for diagnostic tools to assess whether the result can still be trusted.
Most existing diagnostics suffer from one or more of the following weaknesses:
they are specific to a particular setting,
they require evaluating the density of the approximation, which is unavailable for sampling-based methods,
or they are restricted to the marginal distributions of a multivariate posterior.
An overview of diagnostic tools is given in Section~\ref{sec:related_work}.

\begin{figure}
    \centering
    \input{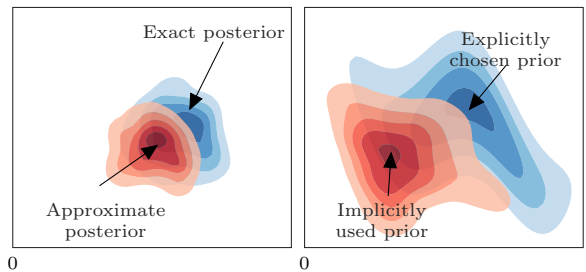}
    \caption{
    \textbf{Left:} a posterior approximation is biased towards solutions of small norm.
    \textbf{Right:}
    the approximation corresponds to the exact posterior under another implicitly defined prior, which is itself biased towards solutions of small norm.
    }
    \label{fig:prior_schematic}
\end{figure}

Existing diagnostics describe the difference to exact Bayesian inference by assessing the mismatch between approximation and true posterior.
In contrast, we investigate a new perspective for diagnostic tools: we describe the approximate inference directly by attributing this mismatch to a change in the inductive bias.
In a fully Bayesian setting, the inductive bias is specified explicitly by the model, which consists of the prior (a priori preference for solutions) and the likelihood (data generating process).
Approximating the posterior can introduce additional bias that is not reflected in the model specification.
We fix the likelihood and only allow the prior to change.
The main idea of this work is to treat the approximation as an exact posterior to the same likelihood and reverse-engineer the corresponding implicitly used prior:
\begin{center}
    \tikzset{>=latex} 

\tikzset{snake it/.style={decorate, decoration=snake}}

\tikzstyle{every node}=[font=\normalsize]
\begin{tikzpicture}[x=3.25in, y=3.25in]
    \node (prior_top) [align=center] {Explicitly\\ chosen prior} ;
    \node (and_top) [right of=prior_top, xshift=8pt] {\&} ;
    \node (like_top) [right of= and_top, xshift=5pt] {Likelihood} ;
    \node (post_top) [right of= like_top, xshift=80pt, align=center, minimum width=63pt] {Exact\\ posterior} ;
    
    \node (prior_bot) [below of=prior_top, yshift=-20pt, align=center] {Implicitly\\ used prior} ;
    \node (and_bot) [below of=and_top, yshift=-20pt] {\&} ;
    \node (like_bot) [below of= like_top, yshift=-20pt] {Likelihood} ;
    \node (post_bot) [below of= post_top, align=center, yshift=-20pt, minimum width=63pt] {Approximate\\ posterior} ;
    
    \draw[->] (like_top) edge node[above] {\textit{exact}} node[below] {\textit{inference}} (post_top) ;
    \draw[->] (like_bot) edge node[above] {\textit{exact}} node[below] {\textit{inference}} (post_bot) ;
    \draw[->, snake it] (post_top) -- (post_bot) ;

    \node (below_prior_bot) [below of = prior_bot, yshift=15pt] {};
    \node (below_post_bot) [below of = post_bot, yshift=15pt] {};
    \draw[dashed, ->] (below_post_bot) to [bend left=15] node [below] {\textit{reverse-engineer}} (below_prior_bot) ;
    
\end{tikzpicture}

\end{center}
This implicit prior describes the inductive bias of the approximation in terms of an a priori preference for solutions.
Figure~\ref{fig:prior_schematic} shows an example of inference based on posterior approximations that are biased towards solutions of small norm. This corresponds to effectively using a different prior with
more mass on solutions of small norm than the explicitly chosen prior.

Let $\left(\like(\cdot|\latent)\right)_{\latent}$ be the likelihood and $\left(\VIapprox(\cdot|\obs)\right)_{\obs}$ the approximations to the posteriors $\left(\posterior(\cdot|\obs)\right)_{\obs}$.
It is reasonable to define the implicit prior to the approximations by fixing an observation $\obs$ and simply reverting Bayes' theorem\footnote{Note that $\yprior$ can be improper, that is, not integrable.} $\yprior(\latent)\propto_\latent \VIapprox(\latent|\obs)/\like(\obs|\latent)$.
Unfortunately, $\yprior$ generally depends on the observation $\obs$.
This means that the approximations to different observations can correspond to different implicit priors, in which case no single distribution $\tilde{\prior}$ satisfies $\VIapprox(\latent|\obs)\propto_\latent \tilde{\prior}(\latent)\like(\obs|\latent)$.
We only have the following weaker interpretation:

\fbox{\parbox{0.97\linewidth}{
Inference based on the approximate posteriors $\left(\VIapprox(\cdot|\obs)\right)_{\obs}$ is exact Bayesian inference with the same likelihood $\left(\like(\cdot|\latent)\right)_{\latent}$, but the prior is chosen from the family $\left(\yprior\right)_\obs$ depending on the observation $\obs$.
}}

Of course, the prior should not depend on the observation if we want to interpret it as the a priori preference for solutions.
To understand the inductive bias of the approximations, we need an observation-independent distribution to compromise between this family of priors.
We look at this problem through the lens of incompatible conditional distributions \citep{Arn:1989}.
This yields a natural solution based on pseudo-Gibbs sampling, which we call the \emph{Gibbs prior}.
An introduction to incompatible conditionals and pseudo-Gibbs sampling is given in Appendix~\ref{app:relwork}.

\paragraph{Observation-(in)dependent diagnostics}
A diagnostic can either treat an approximation under a \emph{fixed observation} $\VIapprox(\cdot|\obs)$ or assess the average behavior of the approximation method \emph{across observations} $\left(\VIapprox(\cdot|\obs)\right)_{\obs}$.
These different tasks can show opposing behavior because an approximation can be good on specific instances but bad in general, or vice versa.
Diagnosing a single approximation helps to understand and improve the inference under the fixed observation, but does not inform about how the approximation method performs in other cases.
In our setting, this task is performed by the distributions $\yprior$.
However, we are interested in the systematic bias of the whole approximation method, which is why we search for an observation-independent compromise between the $\yprior$.
This kind of diagnostic does not guarantee the same behavior on any fixed observation, but helps to understand the method itself.

\paragraph{Contributions}
\begin{itemize}
    \item
    We investigate the novel approach of diagnosing approximate Bayesian inference methods in terms of their inductive bias.
    We show that this requires a compromise and reframe it as a problem of incompatible conditional distributions.
    \item
    We propose the Gibbs prior as a natural solution to the above problem (Section~\ref{sec:main}) and as a diagnostic tool.
    It is based on pseudo-Gibbs sampling, which is widely applicable and easy to implement.
    \item
    We demonstrate how the Gibbs prior can be used to discover the inductive bias of approximate Bayesian inference methods in a Gaussian toy example (Section~\ref{sec:example}) and two intractable Bayesian models (Section~\ref{sec:experiments}).
\end{itemize}

\section{RELATED WORK}\label{sec:related_work}
We divide the literature for diagnostics into two broad categories, depending on how they assess an approximation mismatch.
Diagnostics in the first category compute a divergence between (quantities related to) the posterior and its approximation.
\citet{Gor:2015, Gor:2017} compute Stein discrepancies between the posterior and its approximation.
\citet{Cus:2017} compute the symmetric KL divergence between the approximation and another baseline approximation.
\citet{Dom:2021} computes the symmetric KL divergence between the true joint distribution $p(\obs)\posterior(\latent|\obs)$ and its approximation $p(\obs)\VIapprox(\latent|\obs)$.
\citet{Hug:2020} use the Wasserstein distance to bound the error of posterior point estimates.
Diagnostics in the second category consider derived quantities that are known exactly under the true posterior and test whether they deviate under the approximations.
\citet{Xin:2020} compare a distortion map for posterior cumulative distribution functions to the identity.
\citet{Yu:2021} compare average posterior means and covariances to prior means and covariances.
\citet{Coo:2006} initiate another line of work based on the distribution of posterior quantiles, which is tested for uniformity;
a corrected implemenation is presented by \citet{Tal:2018}.
\citet{Yao:2018} relax the uniformity test of \citet{Coo:2006} and only test for symmetry. They also present another diagnostic based on Pareto-smoothed importance sampling.
\citet{Pra:2014} test for uniformity of $p$-values related to the coverage property; this method is extended by \citet{Rod:2018}.
Our diagnostic also falls into this category where the Gibbs prior is compared to the original prior.
The above diagnostics can also be divided by whether they analyze approximation methods for fixed or general observations.
Our goal of diagnosing average approximation behavior is shared by \citet{Dom:2021, Yu:2021, Coo:2006, Tal:2018, Yao:2018}.

Our diagnostic is based on sampling alternatingly from likelihood and approximation. The same technique was originally used by \citet{Gew:2004} under the name \emph{successive-conditional simulator} with the same goal of diagnosing approximations.
Although both diagnostics are based on the same technique, they apply it differently:
\citet{Gew:2004} uses the simulator without reference to compatibility for generating tuples $(\tilde{\latent}_i, \tilde{\obs}_i)_i$, which are tested against samples from the Bayesian model $(\latent_i,\obs_i)_i$ to assess whether the approximations are exact;
we focus on the marginal values $(\tilde{\latent}_i)_i$ that describe the implicitly used prior to assess the inductive bias.
Our diagnostic is also similar in spirit to \citet{Jos:2020} who link distortions in the likelihood to distortions in the prior.

\section{METHOD}\label{sec:main}

\subsection{Preliminaries}
Let $\prior(\latent)$ be a proper prior distribution on a space of latent variables $\latent\in\Latentspace$ and $\like(\obs|\latent)$ a positive likelihood on a space of observations $\obs\in\Observationspace$.
The corresponding posterior distribution is denoted by $\posterior(\latent|\obs)$.
For every fixed $\obs$ let $\VIapprox(\latent|\obs)$ denote the approximation to the posterior given by the approximate method in question. For sampling-based methods this distribution cannot be evaluated because it is specified only implicitly through samples, which suffices for our diagnostic.
We denote the families of distributions as $\likefam\coloneqq \left(\like(\cdot|\latent)\right)_{\latent\in\Latentspace}$, $\postfam\coloneqq \left(\posterior(\cdot|\obs)\right)_{\obs\in\Observationspace}$, and $\approxfam\coloneqq \left(\VIapprox(\cdot|\obs)\right)_{\obs\in\Observationspace}$.
The families $\likefam$ and $\approxfam$ are called \emph{compatible} if there exists a joint distribution on $\Latentspace\times \Observationspace$ which has $\likefam$ and $\approxfam$ as conditionals.
They are called \emph{incompatible} if they are not compatible \citep{Arn:1989}.

Our goal is to understand the inductive bias of inference based on the approximations $\VIapprox(\latent|\obs)$ in terms of an a priori preference for solutions.
The bias is fully encoded in the original prior $\prior(\latent)$ if the approximation is perfect.
However, a mismatch $\VIapprox(\latent|\obs)\neq\posterior(\latent|\obs)$ can introduce additional bias, which is not captured by the original prior.
The main idea of this paper is to treat the approximation as an exact posterior and look for the corresponding prior distribution $\tilde{\prior}(\latent)$.
This new prior describes the combination of explicitly encoded bias $\prior(\latent)$ and implicitly incurred bias because of approximation mismatch.
We can then compare those priors to gain insights into how the approximation changes the inductive bias.

\subsection{Assessing the Inductive Bias of Posterior Approximations with Gibbs Priors}
This section describes the problem of finding a prior to the approximations from the perspective of incompatible conditionals.
We first motivate the problem by considering fixed observations and then propose a solution based on pseudo-Gibbs sampling.

For a fixed observation $\obs\in\Observationspace$, the implicit \emph{pointwise prior $\yprior$} corresponding to $\VIapprox(\cdot|\obs)$ is defined via
\begin{align}\label{eq:prior_y}
    \yprior(\latent)\propto_\latent \frac{\VIapprox(\latent|\obs)}{\like(\obs|\latent)}\,.
\end{align}
This describes the inductive bias of the approximation $\VIapprox(\cdot|\obs)$ for a fixed observation,
but it is not necessarily the same across different observations.
The pointwise prior $\yprior$ will depend on $\obs$ if and only if the conditional families $\likefam$ and $\approxfam$ are incompatible, which is a simple consequence of the definition.
Informally, the scatter of the family $\ypriorfam$ is an indicator for the degree of compatibility:
in the compatible case, all $\yprior$ are concentrated at some distribution $\yprior\equiv\genprior$,
which is the implicit prior to the approximations.
As the compatibility decreases, $\ypriorfam$ gets more scattered (see Figure~\ref{fig:ex_priors}).
One possible measure of incompatibility is discussed in Appendix~\ref{app:degree}.
As a sanity check, observe that a perfect approximation $\approxfam=\postfam$ recovers the original prior $\prior=\yprior$ for every $\obs$.

Ideally, the inductive bias of approximate inference could be explained by a single prior independent from the observation, like a prior in fully Bayesian inference.
But as the above considerations show, this is not possible if the family $\ypriorfam$ contains different members who offer conflicting explanations.
Therefore, we search for a compromise that reasonably represents the different $\yprior$.
We do so by looking at the situation from the perspective of conditional distributions:
a joint distribution on $\Latentspace\times\Observationspace$ (Bayesian model) is specified indirectly through the conditionals $\likefam$ (likelihood) and $\approxfam$ (posterior approximations). We want to obtain the corresponding $\Latentspace$-marginal (prior).
A standard way to access the joint distribution via its conditionals is Gibbs sampling \citep{Gem:1984, Cas:1992}. Gibbs sampling starts with any initial point $(\latent_0, \obs_0)$ in the joint space and alternatingly updates
$\latent$ given $\obs$ and then $\obs$ given $\latent$.
Under some assumptions, this vector converges to a sample from the joint distribution.
Although Gibbs sampling assumes that the involved conditionals are compatible, it can be used the same way if they are incompatible. In this case it is referred to as \emph{pseudo-Gibbs sampling}, a term coined by \citet{Hec:2001}.
Pseudo-Gibbs sampling leads us to the following candidate prior:
\begin{definition}[Gibbs prior]\label{def:gibbs_prior}
For two families of distributions $\left(\like(\cdot|\latent)\right)_{\latent\in\Latentspace}$ on $\Observationspace$ and $\left(\VIapprox(\cdot|\obs)\right)_{\obs\in\Observationspace}$ on $\Latentspace$ consider the discrete-time Markov chain on $\Latentspace$ whose transition function is given by
\begin{align}\label{eq:transition_fn}
    \transit(\latent^\prime|\latent)
    =\expec_{\Obs\sim \like(\cdot|\latent)}\left[\VIapprox(\latent^\prime|\Obs)\right]\,.
\end{align}
This chain is called the \emph{Gibbs chain}. Any stationary distribution of this Markov chain is called a \emph{Gibbs prior} and denoted by $\Gibbsprior$.
\end{definition}

\begin{figure}
    \centering
    \tikzstyle{every node}=[font=\normalsize]
\begin{tikzcd}[column sep={4.8em,between origins}]
\theta_{1} \arrow[dr, "f\left(\cdot|\theta_{1}\right)"'] \arrow[rr, "r\left(\cdot|\theta_{1}\right)"] & 
& 
\theta_{2} \arrow[dr] & 
& 
\theta_{\infty}\sim\pi_G \\
& 
y_{1} \arrow[ur, "q\left(\cdot|y_{1}\right)"'] & 
& 
\scalebox{2}{$\cdots$} \arrow[ur] & 
\end{tikzcd}
    \caption{Schematic diagram of samples from the Gibbs chain (Definition~\ref{def:gibbs_prior}) with auxiliary variables $\obs_{t}$. The distribution of $\latent_{t}$ converges to the Gibbs prior $\Gibbsprior$.}
    \label{fig:gibbs_schematic}
\end{figure}
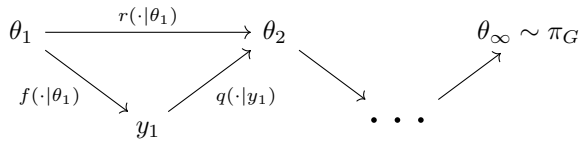

The Gibbs chain is illustrated in Figure~\ref{fig:gibbs_schematic}.
A single step of the chain according to Eq.~\eqref{eq:transition_fn} can be simulated with an  auxiliary variable $\obs$: first sample from the likelihood $\obs\sim \like(\cdot|\latent)$ and then from the approximation $\latent^\prime\sim\VIapprox(\cdot|\obs)$.
Under the caveat of incompatibility, we have the following intuition for the Gibbs prior:

\fbox{\parbox{0.97\linewidth}{
The Gibbs prior describes the a priori preference for solutions of the approximate inference method.
}}

A simple reformulation of the stationarity condition for $\Gibbsprior$ offers two alternative representations
\begin{align}
    \Gibbsprior(\latent)
    &= \int_\Observationspace g(\obs)\VIapprox(\latent|\obs)\diff\obs\label{eq:mixture_q}\\
    &=\int_\Observationspace \tilde{g}(\obs)\like(\obs|\latent)\yprior(\latent)\diff\obs\,,\label{eq:mixture_y}
\end{align}
where $g(\obs)=\int_\Latentspace\Gibbsprior(\tilde{\latent})\like(\obs|\tilde{\latent})\diff\tilde{\latent}$ and $\tilde{g}(\obs)=g(\obs)/\int_\Latentspace\yprior(\tilde{\latent})\like(\obs|\tilde{\latent})\diff\tilde{\latent}$ are weighting functions and Eq.~\eqref{eq:mixture_y} requires all $\yprior$ to be proper.
Eq.~\eqref{eq:mixture_q} shows that the Gibbs prior is a mixture of the pointwise approximations.
This suggests that consistent trends between approximations and posteriors are reflected in the Gibbs prior, for example underestimation of the norm as in Figure~\ref{fig:prior_schematic}.
Eq.~\eqref{eq:mixture_y} relates back to our original motivation of a compromise between $\ypriorfam$ and shows that the Gibbs prior is a mixture of these distributions, reweighted by the likelihood.

\begin{proposition}[Existence and uniqueness of Gibbs priors]\label{prop:gibbs_chain}
Consider two families of distributions $\likefam=\left(\like(\cdot|\latent)\right)_{\latent\in\Latentspace}$ on $\Observationspace$ and $\approxfam=\left(\VIapprox(\cdot|\obs)\right)_{\obs\in\Observationspace}$ on $\Latentspace$. Let $M$ be the corresponding Gibbs chain from Definition~\ref{def:gibbs_prior}.
\begin{enumerate}[label=(\roman*)]
    \item\label{item:gibbs_compatible}
    If $\likefam$ and $\approxfam$ are compatible with joint distribution $p(\latent, \obs)$, then the marginal $p(\latent)$ is a Gibbs prior.
    If $M$ is additionally irreducible, then it is the only Gibbs prior.
    \item\label{item:gibbs_general}
    If $\Latentspace$ and $\Observationspace$ are finite, then there exists a Gibbs prior. If additionally $\likefam$ or $\approxfam$ are positive, then the Gibbs prior is unique.
\end{enumerate}
\end{proposition}
\begin{proof}[Proof (sketch)]
The first statement of part $\ref*{item:gibbs_compatible}$ is a standard Gibbs sampling result; it can be proven by verifying the detailed balance equation for $p(\latent)$, which implies that $M$ is a reversible Markov chain and $p(\latent)$ a stationary distribution.
The statement about uniqueness is trivial, because Gibbs priors are defined as stationary distributions of $M$. A list of sufficient criteria in different settings is given in \citet{Arn:1989}.
Part $\ref*{item:gibbs_general}$ concerns the existence of a (unique) stationary distribution. This condition is a standard result for finite Markov chains, for more general cases see \citet{Nor:1998}.
\end{proof}

Proposition~\ref{prop:gibbs_chain} admits additional interpretations in our Bayesian setting, where $\likefam$ is the likelihood and $\approxfam$ some approximation to the posterior.
Part $\ref*{item:gibbs_compatible}$ states that if $\approxfam$ is the exact posterior under some other prior $\tilde{\prior}$, then this prior is recovered by the Gibbs prior $\Gibbsprior=\tilde{\prior}$.
Part $\ref*{item:gibbs_general}$ shows that Gibbs priors exist under much weaker assumptions than compatibility of $\likefam$ and $\approxfam$.
There are only few other results about the Gibbs chain and its Gibbs priors in the general incompatible case.
\citet{Mur:2019} shows that Gibbs priors are an optimal compromise between incompatible conditionals among a restricted set of distributions.
For discrete distributions, \citet{Kuo:2019} show that the transitions of the Gibbs chain can be interpreted as iterative projections with respect to the KL divergence.

\subsection{Sampling from the Gibbs Prior}\label{sec:algorithm}
\begin{algorithm}
\caption{Simulating the Gibbs chain\protect\footnotemark}
\label{alg:PIGS}
    \KwData{Likelihood $\like$, approximate inference method $\VIapprox$, number of steps $\numsteps$}
    \KwResult{Correlated samples $(\latent_1,\dotsc,\latent_T)$ from $\Gibbsprior$}
        \hbox{$\latent_0 \leftarrow$ \text{Arbitrary initialization, \eg sample from $\prior(\cdot)$}}
        \For{$t\leftarrow 0$ \KwTo $\numsteps-1$}{
        $\phantom{\VIapprox(\cdot|\obs_t)}\mathllap{\obs_t} \leftarrow$ \text{Randomly sample from $\like(\cdot|\latent_t)$}\\
        $\VIapprox(\cdot|\obs_t)\leftarrow$ \text{Approximation to $\posterior(\cdot|\obs_t)$}\\
        $\phantom{\VIapprox(\cdot|\obs_t)}\mathllap{\latent_{t+1}}\leftarrow$ \text{Randomly sample from $\VIapprox(\cdot|\obs_t)$}
        }
\end{algorithm}
\footnotetext{Code available at \href{https://github.com/tml-tuebingen/gibbs-prior-diagnostic}{https://github.com/tml-tuebingen/gibbs-prior-diagnostic}}

Algorithm~\ref{alg:PIGS} describes how to obtain a sequence of correlated samples from the Gibbs prior. Since it is defined as the stationary distribution of the Gibbs chain, this is achieved by simply simulating the chain as in Figure~\ref{fig:gibbs_schematic}.
This approach is very generally applicable because it only requires sampling from the approximate posteriors, but not evaluating their density.
The complexity depends largely on the complexity of computing the approximations to the posterior, which has to be redone every step for a different observation.
The number of steps needed to assure convergence depends on the mixing speed of the Markov chain.
Under the exact posterior, the Gibbs chain mixes fast if there are few observations. Informally, the posterior $\posterior(\latent|\obs)\propto\prior(\latent)\like(\obs|\latent)$ relies heavily on the the prior $\prior$ (the stationary distribution) which ensures that the chain converges to its stationary distribution quickly.
When there are many observations, the posterior concentrates and the high correlation between parameters and observations leads to slow mixing.
In that sense, Algorithm~\ref{alg:PIGS} is more practical under few observations; this case is arguably more interesting because posterior inference gets easier as the number of observations increases.
To ensure that the resulting samples actually correspond to the Gibbs prior, we recommend to monitor convergence of the Gibbs chain \citep{Viv:2020}.

\subsection{How to Use the Gibbs Prior}
There are two principled ways of using the Gibbs prior to diagnose an approximate inference method.
The first way is to assess the quality of the approximation by quantifying the distance to the original prior $\prior$ with some divergence measure $\divergence{\Gibbsprior}{\prior}$, or testing the hypothesis $H_0:\Gibbsprior=\prior$.
A large discrepancy between $\Gibbsprior$ and $\prior$ indicates a bad approximation, because a perfect approximation would yield $\Gibbsprior=\prior$.
The second way is to understand the inductive bias that the approximation imposes by examining the shift in mass from $\prior$ to $\Gibbsprior$.
A direct comparison might not be enlightening if the latent space $\Latentspace$ is large; instead, one could visualize their differences \citep{Llo:2015} or compare the distribution of summary statistics $g\colon\Latentspace\to\R$. 

Note that there are caveats to this interpretation of the Gibbs prior due to incompatibility of likelihood and approximations.
Thinking of the Gibbs prior as the effectively used prior for approximate inference becomes less valid for stronger incompatibility, because the family of pointwise priors $\ypriorfam$ requires a stronger compromise. This is also demonstrated in the next section.

\paragraph{Summary}
We conclude this section by summarizing the three broad cases that can occur when comparing the Gibbs prior $\Gibbsprior$ with the original prior $\prior$:
\begin{enumerate}
    \item\label{case:good} $\Gibbsprior\approx\prior$:
    the Gibbs prior is close to the original prior,
    which suggests that the approximations do not introduce additional bias. In particular, this is the case when the approximations are close to the true posterior. The reverse implication is not necessarily true (Appendix~\ref{app:injective}).
    \item\label{case:bad} $\Gibbsprior\neq\prior$: the Gibbs prior differs from the original prior, which implies that the approximations differ from the true posterior.
    This means that the approximations introduce additional bias, which can be assessed by interpreting the Gibbs prior as the effectively used prior. The validity of this interpretation depends on the compatibility between likelihood and approximations.
    \item\label{case:diverging} The Gibbs chain in Algorithm~\ref{alg:PIGS} does not converge. This can have multiple reasons: the approximations are good but the prior $\prior$ is improper, the approximations are bad, or the chain was not run long enough. 
    We recommend to use the diagnostic conservatively and dismiss it in these cases to avoid falsely rejecting a good approximation.
    To exclude the last case of running the Gibbs chain not long enough, the convergence of the chain should be monitored.
\end{enumerate}

\begin{figure*}[htbp]
\begin{center}
    \begin{subfigure}[b]{.46\linewidth}
    \centering
    \includegraphics{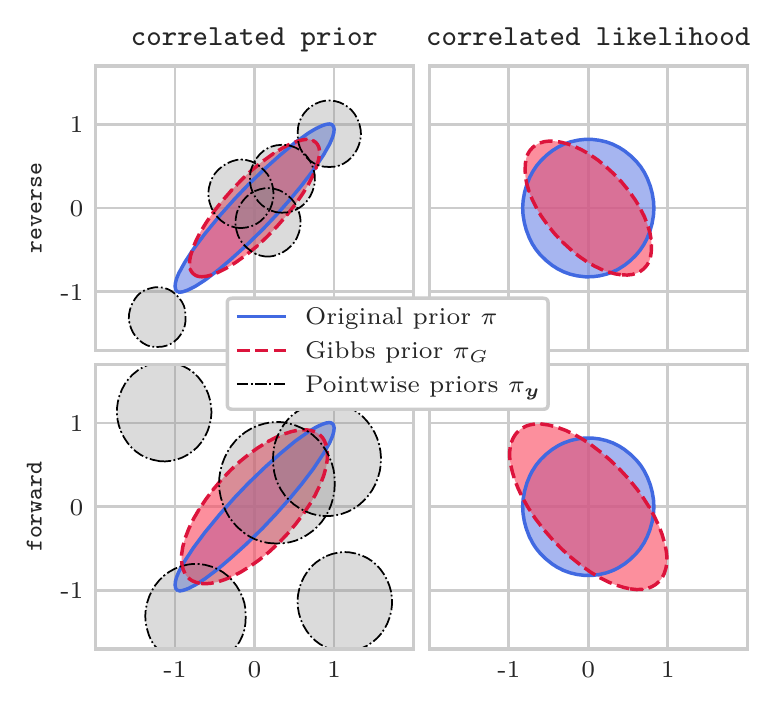}
    \caption{\textbf{Prior distributions.}
    Original prior, Gibbs prior, and pointwise priors for different $\exobs$ (same in both plots).
    }
    \label{fig:ex_priors}
    \end{subfigure}
    \hspace{10pt}
    \begin{subfigure}[b]{.46\linewidth}
    \includegraphics{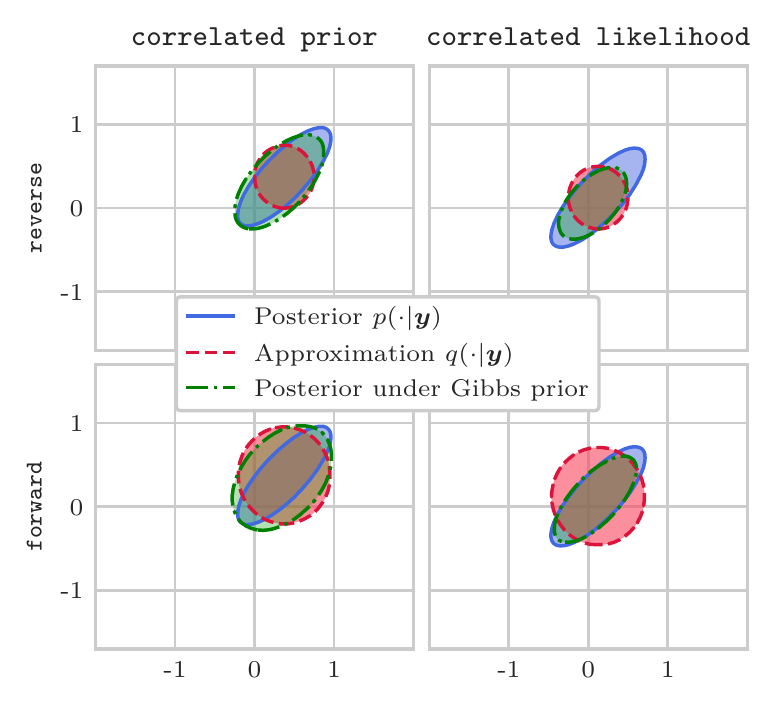}
    \caption{\textbf{Posterior distributions.}
    Posterior, its approximation, and posterior under the Gibbs prior at fixed $\exobs$.
    }\label{fig:ex_posteriors}
    \end{subfigure}
\caption{
Distributions of interest for the variational inference settings described in Section~\ref{sec:toy_model} with $d=2$ and $n=1$.
The setting \setprior\ uses $\CovPrior=I$ and a $\CovLike$ which is strongly correlated along $(1\quad 1)^\top$. For \setlike\ $\CovPrior$ and $\CovLike$ are interchanged. Colored areas show superlevel density sets with mass 0.3.
}
\label{fig:ex_main}
\end{center}
\end{figure*}

\section{ILLUSTRATIVE TOY EXAMPLE}\label{sec:example}
We now give a simple example to demonstrate the concepts from the previous section.
\subsection{Gaussian Toy Model}\label{sec:toy_model}
Consider the problem of estimating the mean $\latent\in\R^d$ of a $d$-dimensional Gaussian distribution with known covariance matrix based on $n$ independent samples $\obs_1,\dotsc,\obs_n\in\R^d$. Placing a Gaussian prior on $\latent$ yields the Bayesian model
\begin{align}\label{eq:model_example}
\begin{split}
    \latent&\overset{\hphantom{\text{indep.}}}{\sim} \Gauss(\MeanPrior, \CovPrior)\,,\\
    \obs_i|\latent &\overset{\text{indep.}}{\sim} \Gauss(\latent, \CovLike)\,, \quad i=1,\dotsc, n\,,
\end{split}
\end{align}
where $\MeanPrior\in\R^d$ and $\CovPrior, \CovLike\in\R^{d\times d}$ are positive definite.
The observations are collected in a matrix $\exobs=(\obs_1,\dotsc,\obs_n)^\top\in\R^{n\times d}$.
We consider four different settings for variational inference in this model,
which are determined by the following two choices:

\paragraph{Correlated posterior}
We choose the prior and likelihood covariance matrices such that the posterior distribution has correlated components. This can be achieved by either a correlated prior and isotropic likelihood (referred to as \setprior) or an isotropic prior and a correlated likelihood (referred to as \setlike).

\paragraph{Variational approximation}
We consider the mean field variational approximation \citep{Bis:2006}.
This method approximates the posterior with the variational family $\MFfamily$, which consists of all distributions on $\R^d$ with independent components. For the objective we consider the commonly used reverse KL divergence
\begin{align}\label{eq:def_inner}
    \VIapprox(\cdot|\exobs) \coloneqq \argmin_{\VIapprox\in\MFfamily}\KL{\VIapprox}{\posterior(\cdot|\exobs)}
\end{align}
(referred to as \setreverse) or the forward KL divergence
\begin{align}\label{eq:def_outer}
    \VIapprox(\cdot|\exobs) \coloneqq \argmin_{\VIapprox\in\MFfamily}\KL{\posterior(\cdot|\exobs)}{\VIapprox}
\end{align}
(referred to as \setforward).

These settings are simple enough so that all distributions of interest are Gaussians and can be computed in closed form. This includes the posteriors $\posterior(\cdot|\exobs)$, the approximations $\VIapprox(\cdot|\exobs)$, the pointwise priors $\prior_{\exobs}$, and the Gibbs prior $\Gibbsprior$. For details see Appendix~\ref{app:example}, which also provides numerical justifications for the following arguments about biases.

\subsection{Bias Discovery Using the Gibbs Prior}\label{sec:toy_bias}
Both approximations \setreverse\ and \setforward\ have two known biases, compactness and loss of correlation \citep{Tur:2011}.
These biases can now also be discovered with the Gibbs prior.
Figure~\ref{fig:ex_priors} shows the priors and Gibbs priors and Figure~\ref{fig:ex_posteriors} shows the corresponding posteriors and approximations.

\paragraph{Bias: compactness}
One known bias of mean field variational inference is the compactness of the approximations as measured by the entropy \citep{Tur:2011}:
comparing the approximations to the true posterior in Figure~\ref{fig:ex_posteriors} shows that they are too compact for \setreverse\ and not compact enough for \setforward.
The same behavior can be observed on the prior level: the Gibbs prior is more compact than the prior for \setreverse\ and less compact for \setforward.

\paragraph{Bias: loss of correlation}
The variational approximations cannot capture any correlation between the coordinates by definition of the variational family $\MFfamily$.
This bias is easily understood on the posterior level, but it is less obvious what this means in terms of an a priori preference for solutions.
In fact, this corresponding preference depends on the source of the posterior correlation and cannot be explained by the posterior alone.
For \setprior, the posterior correlation is caused by the prior correlation.
Uncorrelated approximations therefore correspond to an uncorrelated prior.
The Gibbs priors confirm this intuition by being less correlated than the prior.
For \setlike, the posterior correlation is caused by the likelihood correlation.
Here, the Gibbs priors show that the approximations correspond to a prior whose correlation is orthogonal to the likelihood correlation.
Intuitively, the orthogonal correlations of prior and likelihood ``cancel out'' to produce uncorrelated posteriors.

\subsection{Is the Gibbs Prior a Prior?}
The approximations are exact posteriors under the Gibbs prior if and only if the approximations are compatible to the likelihood.
Equivalently, this is the case when the family of pointwise priors $(\prior_{\exobs})_{\exobs\in\Observationspace}$ concentrates at a single distribution.
Figure~\ref{fig:ex_priors} shows $\prior_{\exobs}$ for various $\exobs$. For \setprior\ they differ strongly and for \setlike\ they are improper and therefore not shown.
In both settings, this implies that the conditionals are incompatible as is typically the case.
This is confirmed by Figure~\ref{fig:ex_posteriors}, which shows that the posteriors under the Gibbs prior do not exactly coincide with the approximations.
Despite these incompatibilities, this example shows that the Gibbs prior can discover inductive biases of the approximate methods.
The Gibbs prior should therefore be thought of as a summary statistic for the inductive bias (see Appendix~\ref{app:gibbs_summary} for more details).

\section{EXPERIMENTS}\label{sec:experiments}
We experiment with the Gibbs prior as a diagnostic tool for various approximations in two Bayesian models. For more details and convergence monitoring of the Gibbs chains see Appendix~\ref{app:exp}.

\paragraph{Baseline} We compare our findings to the diagnostic \citet{Tal:2018}.
This diagnostic is based on the stationarity equation of the prior $\prior$ under the Gibbs chain, but only considers 1-step transitions with some test statistics $f\colon\Theta\to\R$. Under random samples $\tilde{\latent}\sim\prior$, $\tilde{\obs}\sim\like(\cdot|\tilde{\latent})$, and $\latent_1,\dotsc\latent_L\sim\VIapprox(\cdot|\tilde{\obs})$, the rank of $f(\tilde{\latent})$ in $\{f(\latent_1),\dotsc,f(\latent_L)\}$ is computed.
This is repeated over multiple draws of $(\tilde{\latent},\tilde{\obs})$, which gives a histogram of the ranks. Since the histogram is uniform under the exact posterior, any deviations from uniformity indicate an approximation mismatch. We allocate this method the same computational resources in terms of posterior draws as our Gibbs chain.

\subsection{Sum of log-normals}\label{sec:exp_log_normals}
\begin{figure}
    \centering
    \includegraphics{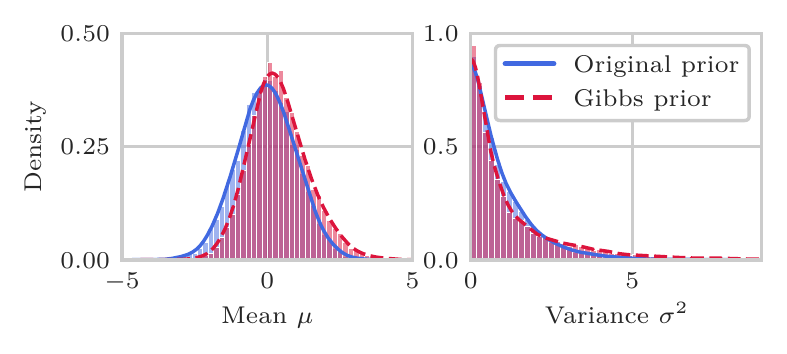}
    \caption{Marginal distributions of prior and Gibbs prior for the sum of log-normals model. A comparison shows that the approximation overestimates $\mu$ and puts more mass on extreme values for $\sigma^2$.}
    \label{fig:abc}
\end{figure}
\begin{figure*}[t]
    \centering
    \includegraphics{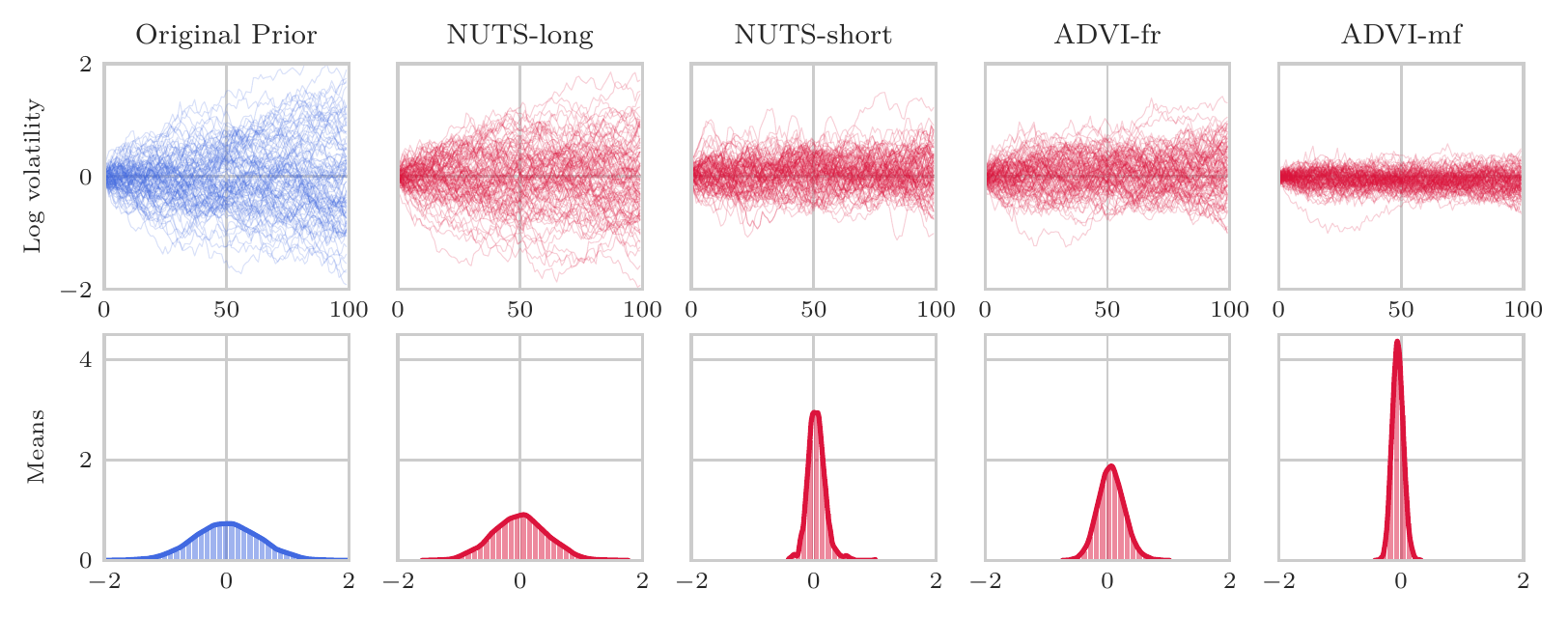}
    \caption{
    \textbf{Top row:} Samples of $\latent\in\R^{100}$ from original prior (blue) and Gibbs priors (red) under various approximations.
    \textbf{Bottom row:} Histograms of the summary statistic $\theta\mapsto1/100\sum_{i=1}^{100}\theta_i$, which is the mean value of a time series. Methods that are closer to the prior introduce less bias.
    }
    \label{fig:volatility}
\end{figure*}
\paragraph{Setup} Our first model describes the sum of $L=10$ independent samples from a log-normal distribution and is given by
\begin{equation*}
    \begin{gathered}
    \mu\sim\Gauss(0, 1),\quad \sigma^2\sim \text{Gamma}(1, 1)\,,\\
    x_l|\latent=(\mu, \sigma^2) \overset{\text{indep.}}{\sim} \LogNormal(\mu, \sigma^2),\quad y=\sum_{l=1}^L x_l\,.
    \end{gathered}
\end{equation*}
Since the corresponding likelihood is infeasible we approximate the posterior in a two-step procedure:
first, we replace the likelihood by its Fenton-Wilkinson approximation \citep{Fen:1960}, which is another log-normal distribution with matching first two moments, and then we use a Laplace approximation to the posterior of this new model.

\paragraph{Bias discovery}
To discover the bias of this approximation we simulate the Gibbs prior based on 10,000 iterations of Algorithm~\ref{alg:PIGS} and show it alongside the original prior in Figure~\ref{fig:abc}.
The first observation is that the Gibbs prior does not coincide with the original prior, which implies that the approximation is not exact.
Furthermore, the deviation between the two distributions is systematic.
For the mean $\mu$, the Gibbs prior has a similar shape as the original prior, but is shifted to the right. 
This implies that the approximations systematically overestimate $\mu$.
For the variance $\sigma^2$, the Gibbs prior puts more mass on extreme values, which means that there is no systematic under- or overestimation.
Compare these findings to \citet{Rod:2018} who consider a fixed approximation to an observation $\obs$ drawn from $\latent=(0,1)$. They confirm that $\mu$ is overestimated, but also find that $\sigma^2$ is underestimated. This does not contradict our findings, because they analyze the approximation to a \emph{fixed} observation, while we analyze the approximations \emph{across} observations.
The other baseline \citet{Tal:2018} is shown in the first two histograms of Figure~\ref{fig:baseline} for the coordinates of $\latent=(\mu,\sigma^2)$ as summary statistics, that is, $f_i(\latent)=\latent_i$.
The histogram for $\mu$ exceeds the confidence region at the smallest rank, which also suggests overestimation.
For $\sigma^2$, the deviation from uniformity is not strong enough to deduce a systematic approximation mismatch.

\subsection{Stochastic Volatility}\label{sec:exp_volatility}
\begin{figure*}[t]
    \centering
    \includegraphics{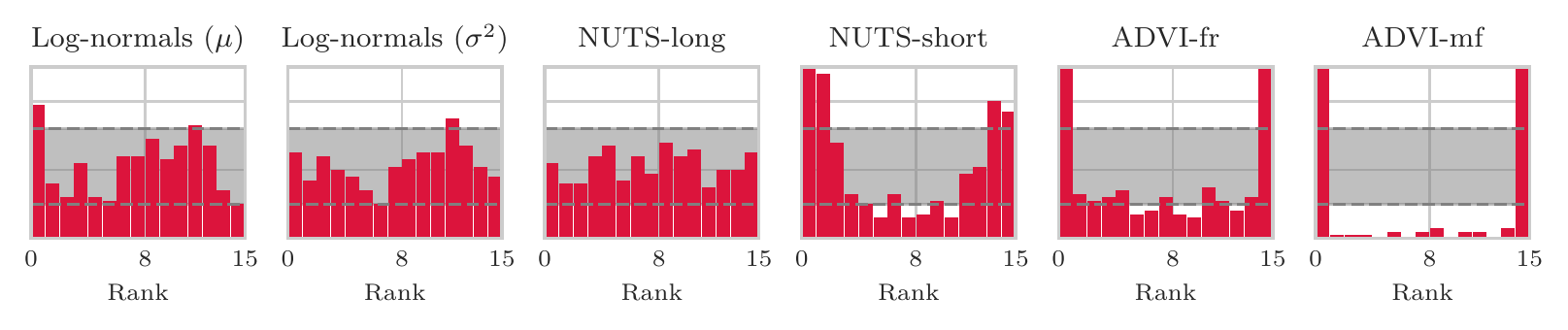}
    \caption{Histograms of rank statistics for the baseline \citet{Tal:2018}. First two histograms are for Section~\ref{sec:exp_log_normals} with coordinates as summary statistics, other histograms are for Section~\ref{sec:exp_volatility} with the mean. Gray band shows a 99\% confidence interval under the exact posterior. Deviations from uniformity indicate approximation mismatch.
    }
    \label{fig:baseline}
\end{figure*}
\paragraph{Setup}
Stochastic volatility models are used in mathematical finance for time series to describe the latent variation of trading price (called the returns).
We consider a model similar to \citet{Hof:2014}:
\begin{align*}
    \theta_i|\theta_{i-1}\sim \Gauss(\theta_i, \sigma^2)&,\quad i=1,\dotsc,T\,,\\
    y_i\overset{\text{indep.}}{\sim}\text{StudentT}(\nu, 0, \exp{\theta_i})&,\quad i=1,\dotsc,T\,,
\end{align*}
where $\theta_0=0, \sigma=.09, \nu=12$, and $T=100$.
The latent parameters $\latent=(\latent_1,\dotsc,\latent_T)$ follow a Gaussian random walk and describe the log volatility of the returns $\obs=(\obs_1,\dotsc,\obs_T)$, which are independent given $\latent$.
As posterior inference methods, we investigate the Hamiltonian Monte Carlo method NUTS \citep{Hof:2014} with different number of steps (10 for NUTS-short and 40 for NUTS-long) and the variational inference method ADVI \citep{Kuc:2017}, which comes in a less powerful mean-field (ADVI-mf) and more powerful full-rank (ADVI-fr) variant.

\paragraph{Bias discovery}
For each approximation method, we can again use the corresponding Gibbs prior in two ways: we test \emph{whether} it deviates from the original prior to assess exactness of the approximation, and if it does, we inspect \emph{how} it deviates to assess the systematic bias.
Figure~\ref{fig:volatility} shows samples from original prior and Gibbs priors under the approximations alongside the distribution of means for each time series as a summary statistic.
Each Gibbs chain was simulated for 10,000 steps, which took 13 hours for ADVI-fr and roughly 5 hours for the other methods on a GPU.
We observe that the Gibbs prior for the long MCMC chain is almost identical to the prior, which confirms that this method is accurate; the Gibbs prior for the corresponding short chain is further away from the prior and closer to the initialization of the chain because it has not fully converged.
The method ADVI-mf shows a strong deviation from the prior by concentrating on less extreme values of the latent variables. This indicates that the approximation is overly compact compared to the true posterior. The same phenomenon was already observed for mean field variational inference in Section~\ref{sec:example}.
It can also be observed for ADVI-fr, but is less pronounced because the method is strictly more powerful.
The baseline \citet{Tal:2018} is shown in the last four histograms of Figure~\ref{fig:baseline} for the same summary statistic as in Figure~\ref{fig:volatility}, the mean value of $\theta$.
For NUTS-long, the histogram stays within the confidence region, which confirms that this method is accurate. The other three methods show a $\cup$-shape, which is most pronounced for ADVI-mf. This indicates that the methods are overly compact and is in line with our findings.
While this baseline can in principle also discover systematic approximation mismatches in terms of over-/underestimation and compactness, the Gibbs prior provides a more complete and nuanced picture.

\section{CONCLUSION AND FUTURE WORK}\label{sec:conclusion}
\paragraph{Conclusion}
We describe a novel diagnostic approach for assessing the inductive bias of approximate Bayesian inference methods. A reformulation of this problem leads to a natural solution, which we call the Gibbs prior. We demonstrate how it can be used to discover the inductive bias in various examples.

\paragraph{Future work}
The Gibbs prior compromises between many pointwise priors. The precise nature of this compromise is intricate, offering several avenues for future analysis. While we introduced the Gibbs prior in the context of approximate Bayesian methods, it can be defined for any generative method returning a distribution over latent variables given an observation.
Another direction is using the pointwise priors as observation-dependent diagnostics. They do not suffer from incompatibility, but can be more challenging to sample from if the approximation density is unknown.

\paragraph{Broader impact}
Recently, there has been a surge of interest in interpretable and explainable machine learning algorithms.
One principled way of explaining an algorithm is to inspect its inductive bias, which describes the preferred solutions independent of the data.
While the inductive bias is specified only implicitly for most algorithms, it is made explicit in Bayesian inference through prior and likelihood.
Unfortunately, this transparency is concealed for approximate Bayesian inference, because approximations introduce additional hidden bias.
We present a method to uncover this inductive bias again, which opens up a new paradigm for the practical evaluation of approximate inference.

\subsubsection*{Acknowledgements}
This work has been supported by the German Research Foundation through the Cluster of Excellence ``Machine Learning -- New Perspectives for Science'' (EXC 2064/1 number
390727645), the BMBF T\"ubingen AI Center (FKZ: 01IS18039A), and the International Max Planck Research School for Intelligent Systems (IMPRS-IS).

\bibliography{main}

\appendix
\onecolumn

\aistatstitle{Discovering Inductive Bias with Gibbs Priors: \\
A Diagnostic Tool for Approximate Bayesian Inference \\
Supplementary Materials}
\aistatsauthor{Luca Rendsburg\affil{1} \And Agustinus Kristiadi\affil{1} \And  Philipp Hennig\affil{1,2} \And Ulrike von Luxburg\affil{1,2}}
\aistatsaddress{%
\affil{1}University of T\"ubingen \hspace{50pt}
\And
\affil{2}Max Planck Institute for Intelligent Systems, T\"ubingen
} 

\section{INCOMPATIBLE CONDITIONALS AND PSEUDO-GIBBS SAMPLING}\label{app:relwork}
When treating approximations as exact posteriors we inevitably face the problem of compatibility, which we shortly introduce in this paragraph.
A bivariate model can be specified explicitly through its joint distribution $p(\latent,\obs)$, for example as in Bayesian models with a marginal $p(\latent)$ (prior) and a conditional distribution $p(\obs|\latent)$ (likelihood).
Alternatively, the model can be specified implicitly through its conditional distributions $p(\latent|\obs)$ and $p(\obs|\latent)$.
Joint modeling simplifies theoretical analysis because closed-form expressions are available, whereas conditional modeling
is less accessible, but more flexible and interpretable.
However, an arbitrary pair of conditional distributions can be \emph{incompatible}, meaning that there exists no joint distribution which produces these conditionals, and if it exists it does not have to be unique \citep{Arn:1989}.
\citet{Arn:2001} argues that ``in general, reasonable-seeming conditional models will not be compatible with any single joint distribution''.
For example, consider the following Bayesian model with real-valued latent variables $\latent$ and observations $\obs$:
for an improper prior $\prior(\latent)= 1$ and a Gaussian likelihood $\like(\obs|\latent)=\Gaussarg{\obs}{\latent}{1}$
Bayes' theorem yields the posterior $\posterior(\latent|\obs)=\Gaussarg{\latent}{\obs}{1}$. Even though both conditional distributions---the likelihood and the posterior---are proper distributions, there exists no proper joint distribution because the corresponding marginal $\prior$ is improper. Hence, the conditionals $\like$ and $\posterior$ are incompatible.
But incompatibility is no all-or-nothing property: even if there exists no joint distribution, one might still look for the joint distribution that is ``most'' compatible with the given conditionals, which leads to notions such as near-compatibility and $\varepsilon$-compatibility \citep{Arn:2002}.
There exist algorithms for assessing the compatibility of conditional distributions \citep{Kuo:2011, Kuo:2017} based on fractions of conditional densities, but most of this theory is restricted to discrete settings.

Gibbs sampling is one of the most natural ways of accessing the joint distribution of a conditionally specified model.
It works by using the conditional distributions to define a time-reversible Markov chain whose stationary distribution is the joint distribution \citep{Has:1970, Gem:1984}.
Gibbs sampling is well-understood and theoretically sound if the conditionals are compatible, but what happens if they are incompatible?
Despite the fact that no joint distribution exists, the Markov chain defined by the conditionals can still converge to a unique stationary distribution, which represents a compromise between the incompatible conditionals \citep{Mur:2019}.
In this case, Gibbs sampling is called pseudo-Gibbs sampling.
Pseudo-Gibbs samplers are widely used, for example in dependency networks \citep{Hec:2001} and missing data imputation \citep{Van:2006, Hug:2014}.
Characterizing the stationary distribution of a pseudo-Gibbs sampler is ongoing research \citep{Shy:2014, Kuo:2019, Mur:2019}.
\clearpage

\section{THE GIBBS PRIOR IS A SUMMARY STATISTIC}\label{app:gibbs_summary}

\subsection{Different Approximations can have the same Gibbs Prior}\label{app:injective}
This section shows that there are fewer Gibbs priors than conditional distributions because different conditionals can define the same Gibbs chain.
For the sake of simplicity, we consider the finite setting with latent space $\Latentspace=[n]=\{1,\dotsc,n\}$ and observation space $\Observationspace=[m]$ with $n,m\in\N$.
Here the likelihood is given by the stochastic matrix $\likefam\in\R^{n\times m}$ and the approximation by another stochastic matrix $\approxfam\in\R^{m\times n}$, that is, $\likefam$ and $\approxfam$ have non-negative entries and their rows sum to 1.
The Gibbs chain from Definition~\ref{def:gibbs_prior} is defined via the transition matrix $P=\likefam \approxfam\in\R^{n\times n}$, which is again a stochastic matrix, and the Gibbs prior is a probability vector $\Gibbsprior\in\R^n$.

The next proposition shows that different approximations $\approxfam\neq\tilde{\approxfam}$ can define the same Gibbs chain. In particular, they define the same Gibbs prior.

\begin{proposition}\label{prop:gibbsmarg_not_injective}
For $n\geq 2$ let $\likefam\in\R^{n\times m},\approxfam\in\R^{m\times n}$ be stochastic matrices with entries in $(0, 1)$ and $\ker\likefam\neq\{0\}$. Then there exists a stochastic matrix $\tilde{\approxfam}\in\R^{m\times n}$ with $Q\neq \tilde{Q}$ that satisfies
\begin{align}\label{eq:same_transmatrix}
    \likefam \approxfam = \likefam\tilde{\approxfam}\,.
\end{align}
In particular, both Markov chains have the same stationary distribution.
\end{proposition}
\begin{proof}
The main idea is to define a suitable perturbation $W$ such that $\tilde{Q}=Q+W$ is a stochastic matrix that satisfies Eq.~\eqref{eq:same_transmatrix}.

Let $0\neq x_0\in\ker F$ and $0\neq w\in\onevec^\perp=\{x\in\R^n~|~x^\top\onevec = 0\}$, where $\onevec\in\R^n$ denotes the vector whose entries are all 1. The vectors $x_0$ and $w$ can be chosen non-zero by the assumptions $\ker F\neq \{0\}$ and $n\geq 2$.
With these vectors, we define the perturbation matrix $W\coloneqq x_0 w^\top\in\R^{m\times n}$ and $\tilde{Q}\coloneqq Q+W\in\R^{m\times n}$. First observe that $Q\neq\tilde{Q}$, because $x_0,w\neq 0$ implies $W\neq 0$.
Using $x_0\in\ker F$, we verify Eq.~\eqref{eq:same_transmatrix} by computing
\begin{align*}
    F\tilde{Q}
    =F(Q+W)
    =FQ + \underbrace{Fx_0}_{=0} w^\top
    =FQ\,.
\end{align*}
It remains to show that $\tilde{Q}$ is a stochastic matrix.
We may assume that $w$ was chosen such that the first condition $0\leq \tilde{Q}=Q+W=Q+x_0w^\top$ holds; otherwise, $w$ can be scaled by an arbitrarily small constant such that this inequality is satisfied, which is always possible because $Q>0$ by assumption.
The other condition is that the rows of $\tilde{Q}$ sum to 1, which we verify with $Q\onevec=\onevec$ (because $Q$ is a stochastic matrix) and $w\in\onevec^\perp$ by computing
\begin{align*}
    \tilde{Q}\onevec = \underbrace{Q\onevec}_{=\onevec} + x_0\underbrace{w^\top\onevec}_{=0} = \onevec\,.
\end{align*}

For the second statement we only need to verify that the stationary distribution of the Markov chain defined with the transition matrix $P=\likefam\approxfam$ indeed uniquely exists. This is the case because the assumptions $\likefam,\approxfam>0$ imply $P>0$, hence the corresponding Markov chain is positive recurrent with finite state space. This implies the existence of a unique stationary distribution.
\end{proof}

\begin{example}
An example for Proposition~\ref{prop:gibbsmarg_not_injective} with $n=2$ and $m=3$ is given by the matrices
\begin{align*}
    F=
    \begin{pmatrix}
    .1 & .4 & .5\\
    .3 & .2 & .5
    \end{pmatrix}\,,
    \quad
    Q=
    \begin{pmatrix}
    .2 & .8 \\
    .4 & .6 \\
    .5 & .5
    \end{pmatrix}\,,
    \quad
    \tilde{Q}=
    \begin{pmatrix}
    .1 & .9 \\
    .3 & .7 \\
    .6 & .4
    \end{pmatrix}\,,
\end{align*}
which satisfy $Q\neq \tilde{Q}$ and
\begin{align*}
    FQ =
    \begin{pmatrix}
    .43 & .57\\
    .39 & .61
    \end{pmatrix}
    = F\tilde{Q}\,.
\end{align*}
\end{example}

\subsection{A Weaker Notion of Compatibility between Conditional Distributions is Sufficient}\label{app:remedy}
In this section, we argue that the notion of compatibility between conditional distributions is actually stricter than necessary for assessing whether the Gibbs prior provides a useful explanation.
We do so by introducing a weaker notion of compatibility under which the Gibbs prior retains a strong interpretation.
First, we recap the setting as presented in the main paper.
For a Bayesian model with likelihood $\likefam$ we are given approximations $\approxfam$ to the true posterior, and our goal is to assess their inductive bias in terms of an a priori preference for solutions.
We propose to consider another fully Bayesian model $\mathcal{M}_G$, specified with the same likelihood $\likefam$ and the Gibbs prior $\Gibbsprior$.
The Gibbs prior $\Gibbsprior$ can then be used to reason about the inductive bias of $\approxfam$ if the conditional distributions $\likefam$ and $\approxfam$ are compatible, because then the posteriors under $\mathcal{M}_G$ coincide with $\approxfam$.

However, even when they are different, the Bayesian model $\mathcal{M}_G$ can accurately describe inference based on $\approxfam$.
This is achieved by considering the whole pipeline of inference instead of inference based on a fixed observation:
starting with an unknown true latent parameter $\latent$, we observe some data through the likelihood $\obs\sim\like(\cdot|\latent)$, based on which we use the approximations to estimate the latent parameter $\latent^\prime\sim\VIapprox(\cdot|\obs)$.
This process is summarized in the probabilities of estimating $\latent^\prime$ if the true parameter is $\latent$, which are precisely the transition probabilities of the Gibbs chain.
Defining the same Gibbs chain as the approximations is therefore sufficient for the Bayesian model $\mathcal{M}_G$ to qualify as an interpretable reformulation. This leads to the following weaker notion of compatibility between conditional distributions:
\begin{definition}[Weak compatibility of conditional distributions]\label{def:weak_compatibility}
For two families of conditional distributions $\likefam=\left(\like(\cdot|\latent)\right)_{\latent\in\Latentspace}$ on $\Observationspace$ and $\approxfam=\left(\VIapprox(\cdot|\obs)\right)_{\obs\in\Observationspace}$ on $\Latentspace$, let $\Gibbsprior$ denote the corresponding Gibbs prior from Definition~\ref{def:gibbs_prior}.
Let $\Gibbspostfam=\left(\Gibbspost(\cdot|\obs)\right)_{\obs\in\Observationspace}$ denote the posteriors under to the Bayesian model specified by $\Gibbsprior$ and $\likefam$.
Then $\likefam$ and $\approxfam$ are called \emph{weakly compatible}, if the Gibbs chain of $\likefam$ and $\approxfam$ coincides with the Gibbs chain of $\likefam$ and $\Gibbspostfam$, that is,
\begin{alignat*}{3}
    &&
    \expec_{\Obs\sim \like(\cdot|\latent)}\left[\VIapprox(\latent^\prime|\Obs)\right]
    =\expec_{\Obs\sim \like(\cdot|\latent)}\left[\Gibbspost(\latent^\prime|\Obs)\right]
    &&\quad
    \forall \latent\in\Latentspace\,.
\end{alignat*}
\end{definition}

As the naming suggests, weak compatibility of two conditional distributions is strictly weaker than compatibility: compatibility trivially implies weak compatibility, whereas Proposition~\ref{prop:gibbsmarg_not_injective} shows that the converse is not true.
Since the Gibbs prior can be used to reason about conditionals that are only weakly compatible, this means that it is useful in more situations than what compatibility suggests.
In particular, there exist different conditional distributions $\approxfam$ which justifiably get assigned the same Gibbs prior, because they yield the same Gibbs chain.

\section{MEASURING THE DEGREE OF COMPATIBILITY}\label{app:degree}
Most existing literature focuses on the question whether families of conditional distributions are exactly compatible \citep{Kuo:2011, Kuo:2017}.
However, in the context of this paper, the more relevant question is how incompatible they are, requiring a practical way to measure the degree of compatibility.
This leads to notions such as near-compatibility and $\varepsilon$-compatibility \citep{Arn:2002} and is based on computing some distance between joint distributions, which involve the conditional distributions \citep{Gho:2015}.

In this section, we present a practical way of measuring the degree of compatibility between likelihood and approximations.
Recall that the Gibbs chain from Definition~\ref{def:gibbs_prior} is based on alternate sampling from likelihood and approximation as depicted in Figure~\ref{fig:gibbs_schematic}. A sequence of this chain has the form $\left(\latent_{1}, \obs_{1}, \latent_{2}, \obs_{1},\dotsc\right)$ with $\latent_{t}\in\Latentspace$ and $\obs_{t}\in\Observationspace$.
We then defined the Gibbs prior $\Gibbsprior$ as the limiting distribution of the $\latent_{t}$, but we can analogously consider the limiting distribution of the $\Observationspace$-components $\obs_{t}$. To this end, we generalize Definition~\ref{def:gibbs_prior} by looking viewing the Gibbs chain as a Markov chain on $\Latentspace\times\Observationspace$:
\begin{definition}[Gibbs chain (extension to Definition~\ref{def:gibbs_prior})]\label{def:gibbs_prior_extended}
For two families of distributions $\left(\like(\cdot|\latent)\right)_{\latent\in\Latentspace}$ on $\Observationspace$ and $\left(\VIapprox(\cdot|\obs)\right)_{\obs\in\Observationspace}$ on $\Latentspace$ consider the discrete-time Markov chain on $\Latentspace\times\Observationspace$ whose transition function is given by
\begin{align*}
    \transit((\latent^\prime, \obs^\prime)|(\latent, \obs))
    =\like(\obs^\prime|\latent)\VIapprox(\latent^\prime|\obs^\prime)\,.
\end{align*}
This chain is called the \emph{Gibbs chain}.
The projection onto the $\Latentspace$-components is a Markov chain on $\Latentspace$, any stationary distribution of which is called a \emph{Gibbs prior} and denoted by $\Gibbsprior$.
The projection onto the $\Observationspace$-components is a Markov chain on $\Observationspace$, any stationary distribution of which is and denoted by $p_G$.
\end{definition}
\citet{Kuo:2019} also studied this Gibbs chain for discrete distributions.
In the main paper we specified a joint distribution on $\Latentspace\times\Observationspace$ with the Gibbs prior $\Gibbsprior$ as the $\Latentspace$-marginal and the likelihood $\likefam$ as the corresponding conditional.
The main observation for measuring the degree of compatibility between $\likefam$ and $\approxfam$ is that we can also specify a joint distribution from the other direction, that is, with $p_G$ as the $\Observationspace$-marginal and $\approxfam$ as the corresponding conditional.
We abbreviate those two joint distributions with $\Gibbsjoint$ and $\Evidencejoint$.
They coincide if and only if $\likefam$ and $\approxfam$ are compatible.
It is therefore natural to measure the degree of compatibility via some divergence between them.

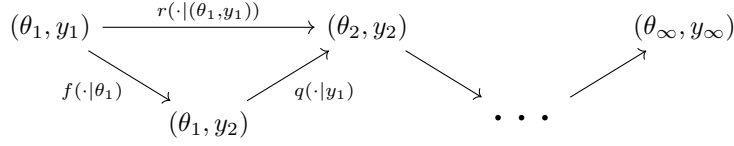
\begin{figure}
    \centering
    \tikzstyle{every node}=[font=\normalsize]
\begin{tikzcd}[column sep={6em,between origins}]
\left(\theta_{1},y_{1}\right) \arrow[dr, "f\left(\cdot|\theta_{1}\right)"'] \arrow[rr, "{r\left(\cdot|\left(\theta_{1} ,y_{1}\right)\right)}"] & 
& 
\left(\theta_{2},y_{2}\right) \arrow[dr] & 
& 
\left(\theta_{\infty},y_{\infty}\right) \\
& 
\left(\theta_{1},y_{2}\right) \arrow[ur, "q\left(\cdot|y_{1}\right)"'] & 
& 
\scalebox{2}{$\cdots$} \arrow[ur] & 
\end{tikzcd}
    \caption{Schematic diagram of samples from the Gibbs chain from Definition~\ref{def:gibbs_prior_extended}, where one step first updates $\obs$ with $\like$ and then $\latent$ with $\VIapprox$. The distribution of $\latent_{t}$ converges to the Gibbs prior $\Gibbsprior$ and the distribution of $\obs_{t}$ converges to $p_G$.}
    \label{fig:gibbs_schematic_extension}
\end{figure}

\paragraph{A practical algorithm}
We can obtain (correlated) samples from the two joint distributions $\Gibbsjoint$ and $\Evidencejoint$ with the same Gibbs chain used for obtaining samples from the Gibbs prior $\Gibbsprior$. Simulating the Gibbs chain is described by Figure~\ref{fig:gibbs_schematic_extension}. It yields a sequence $\left(\latent_{1}, \obs_{1}, \latent_{2}, \obs_{1},\dotsc\right)$, where the marginal distributions of $\latent_{t}$ and $\obs_{t}$ converge to $\Gibbsprior$ and $p_G$, respectively. Since the components are updated alternatingly with the conditional distributions, we can pair the entries to obtain samples from the joint distributions. However, the order of the pairing is important:
\begin{itemize}
    \item $(\latent_{t}, \obs_{t})\sim \Gibbsjoint$, because $\latent_{t}\sim\Gibbsprior$ (for large $t$) and $\obs_{t}\sim\like(\cdot|\latent_{t})$
    \item $(\latent_{t+1}, \obs_{t})\sim \Evidencejoint$, because $\obs_{t}\sim p_G$ (for large $t$) and $\latent_{t+1}\sim\VIapprox(\cdot|\obs_{t})$
\end{itemize}
This leads to basically the same algorithm as Algorithm~\ref{alg:PIGS} in the main paper, except that the auxiliary variables $\obs_{t}$ are stored as well and paired accordingly with no computational overhead:

\begin{algorithm}
\caption{
Simulating the Gibbs chain (samples from the joint distributions $\Gibbsjoint$ and $\Evidencejoint$)
}
    \KwData{Likelihood $\like$, approximate inference method $\VIapprox$, number of steps $\numsteps$}
    \KwResult{Correlated samples $(\latent_t,\obs_t)_{t=0}^{T-1}$ from $\Gibbsprior$ and $(\latent_{t+1},\obs_t)_{t=0}^{T-1}$ from $p_G$}
        \hbox{$\latent_0 \leftarrow$ \text{Arbitrary initialization, \eg sample from $\prior(\cdot)$}}
        \For{$t\leftarrow 0$ \KwTo $\numsteps-1$}{
        $\phantom{\VIapprox(\cdot|\obs_t)}\mathllap{\obs_t} \leftarrow$ \text{Randomly sample from $\like(\cdot|\latent_t)$}\\
        $\VIapprox(\cdot|\obs_t)\leftarrow$ \text{Approximation to $\posterior(\cdot|\obs_t)$}\\
        $\phantom{\VIapprox(\cdot|\obs_t)}\mathllap{\latent_{t+1}}\leftarrow$ \text{Randomly sample from $\VIapprox(\cdot|\obs_t)$}
        }
    \label{alg:PIGS_joint}
\end{algorithm}

\begin{figure}[t]
\begin{center}
    \begin{subfigure}[b]{.46\linewidth}
    \centering
    \includegraphics{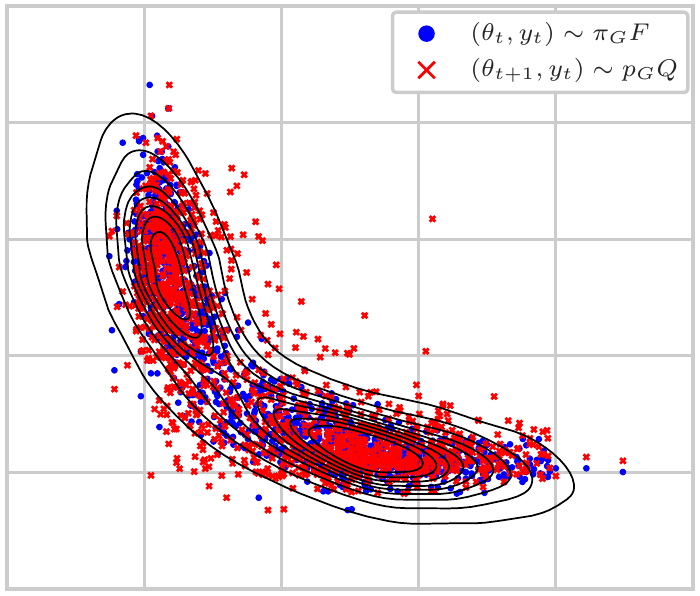}
    \caption{
    Correlated samples from $\Gibbsjoint$ and $\Evidencejoint$ for two compatible conditionals with the underlying joint distribution as contours. The conditionals being compatible is equivalent to all three joint distributions coinciding.
    }
    \label{fig:compatible}
    \end{subfigure}
    \hspace{10pt}
    \begin{subfigure}[b]{.46\linewidth}
    \includegraphics{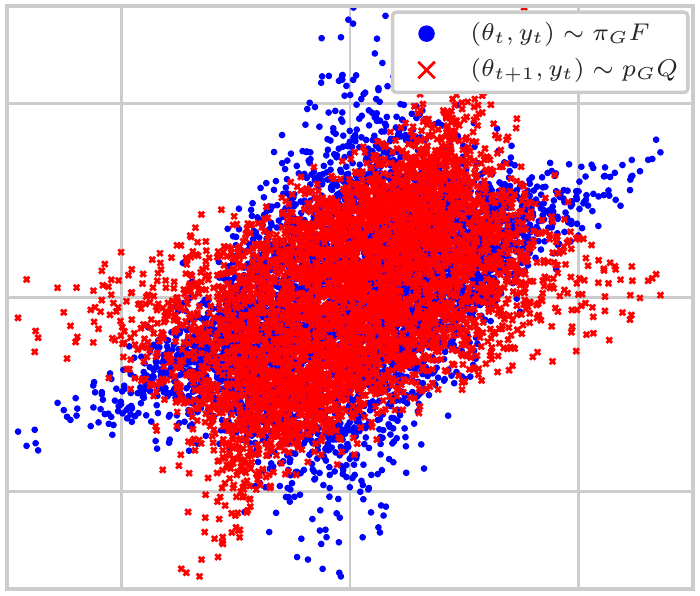}
    \caption{
    Correlated samples from $\Gibbsjoint$ and $\Evidencejoint$ for two incompatible conditionals.
    The conditionals being incompatible is equivalent to those two joint distributions being different.
    }
    \label{fig:incompatible}
    \end{subfigure}
\caption{Samples from Algorithm~\ref{alg:PIGS_joint} for the Gaussian conditional distributions of Example~\ref{ex:compatibility}.}
\label{fig:examples_compatibility}
\end{center}
\end{figure}

\begin{example}[Gaussian conditional distributions]\label{ex:compatibility}
We demonstrate Algorithm~\ref{alg:PIGS_joint} for two pairs of Gaussian conditional distributions with $\Latentspace=\Observationspace=\R$, one compatible and one incompatible.
The first example is taken from \citet{Arn:2001} and given by the conditional distributions
\begin{align*}
    \like(\obs|\latent) = \Gaussarg{\obs}{\frac{4}{1+\latent^2}}{\frac{1}{1+\latent^2}}
    \qand
    \VIapprox(\latent|\obs) = \Gaussarg{\latent}{\frac{4}{1+\obs^2}}{\frac{1}{1+\obs^2}}\,.
\end{align*}
These conditionals are compatible with the bivariate joint density
\begin{align*}
    p(\latent, \obs) = \exp\left((1,\; \latent,\; \latent^2)
    \begin{pmatrix}
    c & 4 & -1/2 \\
    4 & 0 & 0 \\
    -1/2 & 0 & -1/2
    \end{pmatrix}
    \begin{pmatrix}
    1 \\
    \obs \\
    \obs^2
    \end{pmatrix}
    \right)\,,
\end{align*}
where $c\in\R$ plays the role of the normalizing constant.
The corresponding samples from Algorithm~\ref{alg:PIGS_joint} are shown in Figure~\ref{fig:compatible} together with a contour plot of the true joint density $p(\latent,\obs)$. All three joint densities $\Gibbsjoint$, $\Evidencejoint$, and $p$ overlap, confirming the compatibility of these conditional distributions.

The other example is given by the conditional densities
\begin{align*}
    \like(\obs|\latent) = \Gaussarg{\obs}{\frac{\latent}{2}}{\frac{1}{1+\latent^2}}
    \qand
    \VIapprox(\latent|\obs) = \Gaussarg{\latent}{\frac{\obs}{2}}{\frac{1}{1+\obs^2}}\,.
\end{align*}
These conditional distributions are incompatible (\citet{Arn:2001} gives a full characterization of compatible Gaussian conditional distributions). Therefore the joint distributions $\Gibbsjoint$ and $\Evidencejoint$ cannot coincide exactly. This is confirmed by Figure~\ref{fig:incompatible}, which shows samples from the two joint distributions. Based on these samples we could now measure some kind of divergence between the two distributions to assess the degree of compatibility.
\end{example}

\section{RESULTS AND PROOFS FOR THE GAUSSIAN TOY EXAMPLE}\label{app:example}

\subsection{Distributions in the Gaussian Toy Example}
This section computes the distributions of interest for the Bayesian model defined in Eq.~\eqref{eq:model_example} from Section~\ref{sec:example}. This includes the posterior, the approximations, and the pointwise prior in Proposition~\ref{prop:distributions_ex}, as well as the Gibbs prior in Theorem~\ref{thm:gibbsmarg_ex}.

\begin{proposition}[Posterior, approximations, and pointwise priors]\label{prop:distributions_ex}
Consider the Bayesian model defined in Eq.~\eqref{eq:model_example} and let $\latent\in\Latentspace$ and $\exobs=(\obs_1,\dotsc,\obs_n)\in\Observationspace$.
\begin{enumerate}[label=(\roman*)]
    \item\label{item:post} The posterior distribution is given by
    \begin{align}\label{eq:ex_posterior}
        \posterior(\latent|\exobs) = \Gaussarg{\latent}{\MeanPost}{\CovPost}\,,
    \end{align}
    where $\CovPost = \left(\CovPrior^{-1} + n\CovLike^{-1}\right)^{-1}$, $\MeanPost=\CovPost\left(\CovPrior^{-1}\MeanPrior + n\CovLike^{-1}\avgobs\right)$, and $\avgobs=1/n\sum_{j=1}^n\obs_j$.
    \item\label{item:approx}
    The mean field variational approximation is given by
    \begin{align}\label{eq:ex_approx}
        \VIapprox(\latent|\exobs) = \Gaussarg{\latent}{\MeanPost}{\CovApprox}\,,
    \end{align}
    where
    \begin{align}\label{eq:ex_CovApprox}
        \CovApprox =
        \begin{cases}
            \diag\left(\CovPost^{-1}\right)^{-1},&\text{for $\VIapprox$ defined via Eq.~\eqref{eq:def_inner}}\textnormal{ (\setreverse)}\\
            \diag\left(\CovPost\right),&\text{for $\VIapprox$ defined via Eq.~\eqref{eq:def_outer}\textnormal{ (\setforward)}}
        \end{cases}\,.
    \end{align}
    Hereby, the $\diag$ operator keeps the diagonal entries of a matrix and sets all off-diagonal entries to 0.
    \item\label{item:y_marg} Whether the pointwise prior $\ypriorex$ is a proper distribution depends on the matrix $\CovApprox^{-1}-n\CovLike^{-1}$. If it is positive definite, then
    \begin{align*}
        \ypriorex(\latent)\propto\Gaussarg{\latent}{\Meany}{\Covy}\,,
    \end{align*}
    where $\Covy=\left(\CovApprox^{-1}-n\CovLike^{-1}\right)^{-1}$ and $\Meany=\Covy\left(\CovApprox^{-1}\MeanPost - n\CovLike^{-1}\avgobs\right)$. Otherwise, $\ypriorex$ is improper. In particular, $\ypriorex$ is always proper in the setting \textnormal{\setprior}.
\end{enumerate}
\end{proposition}

\begin{theorem}[Gibbs prior]\label{thm:gibbsmarg_ex}
The Gibbs marginal $\Gibbsprior$ to the Bayesian model defined in Eq.~\eqref{eq:model_example} is given by
\begin{align}\label{eq:gibbsmarg_ex}
    \Gibbsprior(\latent)=\Gaussarg{\latent}{\MeanPrior}{\CovGibbs}\,.
\end{align}
Hereby, $\MeanPrior$ is the mean of the prior distribution $\prior$ and $\CovGibbs$ satisfies the Lyapunov equation
\begin{align}\label{eq:Lyapunov}
    A\CovGibbs A^\top - \CovGibbs + B = 0\,
\end{align}
where $A=n\CovPost\CovLike^{-1}$ and $B=\CovApprox + n\CovPost\CovLike^{-1}\CovPost$ with $\CovPost$ and $\CovApprox$ defined as in Proposition~\ref{prop:distributions_ex}.
\end{theorem}

\begin{proof}[Proof of Proposition~\ref{prop:distributions_ex}]
Recall the density function of a multivariate normal distribution $\Gaussarg{\latent}{\mu}{\Sigma}$, which is given by
\begin{align*}
    \Gaussarg{\latent}{\mu}{\Sigma}
    =(2\pi)^{-\frac{d}{2}}\det(\Sigma)^{-\frac{1}{2}}\exp\left(-\frac{1}{2}(\latent-\mu)^\top\Sigma^{-1}(\latent-\mu)\right)
    \propto_\latent \exp\left(-\frac{1}{2}\left[\latent^\top\Sigma^{-1}\latent - 2\latent^\top\Sigma^{-1}\mu\right]\right)\,.
\end{align*}

{\bf Proof of $\ref*{item:post}$.}
First observe that up to proportionality the likelihood $\like(\exobs|\latent)$ as a function of $\latent$ depends only on the average observation $\avgobs=1/n\sum_{j=1}^n\obs_j$
\begin{align}\label{eq:like_propto_avg}
    \like(\exobs|\latent)
    =\prod_{j=1}^n\Gaussarg{\obs_j}{\latent}{\CovLike}
    &\propto_\latent \prod_{j=1}^n \exp\left(-\frac{1}{2}\left[\latent^\top\CovLike^{-1}\latent - 2\latent^\top\CovLike^{-1}\obs_j\right]\right)\notag\\
    &= \exp\left(-\frac{1}{2}\left[\latent^\top n\CovLike^{-1}\latent - 2\latent^\top n\CovLike^{-1}\avgobs\right]\right)\notag\\
    &\propto_\latent\Gaussarg{\avgobs}{\latent}{1/n\CovLike}\,.
\end{align}
Together with $\prior(\latent)=\Gaussarg{\latent}{\MeanPrior}{\CovPrior}$, Bayes' theorem yields
\begin{align*}
    \posterior(\latent|\exobs)
    \propto_\latent\prior(\latent)\like(\exobs|\latent)
    &\propto_\latent \exp\left(-\frac{1}{2}\left[\latent^\top\CovPrior^{-1}\latent - 2\latent^\top\CovPrior^{-1}\MeanPrior\right]\right)\exp\left(-\frac{1}{2}\left[\latent^\top n\CovLike^{-1}\latent - 2\latent^\top n\CovLike^{-1}\avgobs\right]\right)\\
    &=\exp\left(-\frac{1}{2}\left[\latent^\top\left(\CovPrior^{-1}+n\CovLike^{-1}\right)\latent - 2 \latent^\top\left(\CovPrior^{-1}\MeanPrior + n\CovLike^{-1}\avgobs\right)\right]\right)\\
    &=\exp\left(-\frac{1}{2}\left[\latent^\top\CovPost^{-1}\latent - 2 \latent^\top\CovPost^{-1}\MeanPost\right]\right)\\
    &\propto \Gaussarg{\latent}{\MeanPost}{\CovPost}\,.
\end{align*}
Note that $\CovPrior^{-1} + n\CovLike^{-1}$ is positive definite as the sum of two positive definite matrices, and therefore $\CovPost=\left(\CovPrior^{-1} + n\CovLike^{-1}\right)^{-1}$ is positive definite as well.

{\bf Proof of $\ref*{item:approx}$.}
By definition of the mean-field variational family, every variational density factorizes as $q(\theta|\exobs)=\prod_{j=1}^m q_j(\theta_j)$.
For the setting \setforward\, we refer to \citep[Section 10.1.2]{Bis:2006}, where it is shown that the optimal $q_j$ simply coincide with the marginal densities of the posterior $q_j(\theta_j)=p(\theta_j|\exobs)=\Gaussarg{\theta_j}{(\MeanPost)_j}{\left(\CovPost\right)_{j,j}}$.

For the other setting \setreverse\ let $\expec_{-j}$ denote the expectation over all latent variables $\theta_i$ except $\theta_j$ with respect to the factorized distribution $\prod_{i\neq j}q_i(\theta_i)$. To simplify the following computation, we abbreviate $\Sigma_{n}^{-1}\eqqcolon\Gamma$ and $\mu_n\eqqcolon\mu$ (now $\mu_k$ refers to the $k$-th component of $\MeanPost$).
We use that the optimal solution satisfies the recursive update rule
\begin{align*}
    q_j(\theta_j)
    &\propto_{\theta_j} \exp\left(\expec_{-j}\log p(\theta,\exobs)\right)\tag{\citet{Bis:2006}}\\
    &\propto_{\theta_j} \exp\left(\expec_{-j}\log p(\theta|\exobs)\right)\\
    &\propto_{\theta_j} \exp\left(-\frac{1}{2}\expec_{-j}\left[\left(\theta-\mu\right)^\top\Gamma\left(\theta-\mu\right)\right]\right)\tag{$p(\theta|\exobs)=\Gaussarg{\theta}{\mu}{\Gamma^{-1}}$ by Eq.~\eqref{eq:ex_posterior}}\\
    &\propto_{\theta_j}\exp\left(-\frac{1}{2}\left(\Gamma_{j,j}
    \left(\theta_j-\mu_j\right)^2 + 2\left(\theta_j-\mu_j\right)\sum_{k\neq j}\Gamma_{j,k}\left(m_k-\mu_k\right)\right)\right)\tag{$m_k\coloneqq\expec_{q_k}\latent_k$}\\
    &\propto_{\theta_j}\exp\left(-\frac{1}{2\Gamma_{j,j}^{-1}
    }\left(\theta_j-\mu_j+\frac{1}{\Gamma_{j,j}}\sum_{k\neq j}\Gamma_{j,k}(m_k-\mu_k)\right)^2\right)\\
    &\propto_{\latent_j}\Gaussarg{\latent_j}{\mu_j-\frac{1}{\Gamma_{j,j}}\sum_{k\neq j}\Gamma_{j,k}(m_k-\mu_k)}{\Gamma_{j,j}^{-1}}\\
    &=\Gaussarg{\latent_j}{m_j}{\Gamma_{j,j}^{-1}}\tag{Definition of $m_j$}\,.
\end{align*}
This shows that the solutions $q_j$ are normally distributed and have the claimed variance $\Gamma_{j,j}^{-1}=\left(\CovPost^{-1}\right)_{j,j}^{-1}$. However, their means $m_j$ are only recursively determined and to conclude the proof, we need to show that $m_j=\mu_j$. The last equation in the previous computation gives the recursive relation
\begin{alignat*}{2}
    && m_j &= \mu_j-\frac{1}{\Gamma_{j,j}}\sum_{k\neq j}\Gamma_{j,k}(m_k-\mu_k) \hspace{50pt}\forall j=1,\dotsc,m\\
    \Leftrightarrow\hspace{30pt} && \frac{1}{\Gamma_{j,j}}\sum_{k=1}^m\Gamma_{j,k}m_k &= \frac{1}{\Gamma_{j,j}}\sum_{k=1}^m\Gamma_{j,k}\mu_k \hspace{30pt}\forall j=1,\dotsc,m\\
    \Leftrightarrow\hspace{50pt} && \scalprod{\Gamma_j}{\bo{m}} &= \scalprod{\Gamma_j}{\mu} \hspace{30pt}\forall j=1,\dotsc,m\\
    \Leftrightarrow\hspace{50pt} && \Gamma\bo{m} &= \Gamma\mu\,,
\end{alignat*}
where $\Gamma_j$ denotes the $j$-th row of $\Gamma$ and $\bo{m}$ is the vector containing all $m_k$. Since $\Gamma$ is positive definite, the last equality implies $\bo{m}=\mu$ and concludes the proof.

{\bf Proof of $\ref*{item:y_marg}$.}
We can compute the pointwise prior $\ypriorex(\latent)$ with its definition in Eq.~\eqref{eq:prior_y} with Eq.~\eqref{eq:ex_approx} for $\VIapprox(\latent|\exobs)$ and Eq.~\eqref{eq:like_propto_avg} for $\like(\exobs|\latent)$ as
\begin{align}\label{eq:is_yprior_proper}
    \ypriorex(\latent)
    \propto_\latent\frac{\VIapprox(\latent|\exobs)}{\like(\exobs|\latent)}
    \propto_\latent\frac{\Gaussarg{\latent}{\MeanPost}{\CovApprox}}{\Gaussarg{\avgobs}{\latent}{1/n\CovLike}}
    &\propto_\latent \frac{\exp\left(-\frac{1}{2}\left[\latent^\top\CovApprox^{-1}\latent - 2\latent^\top\CovApprox^{-1}\MeanPost\right]\right)}{\exp\left(-\frac{1}{2}\left[\latent^\top n\CovLike^{-1}\latent - 2\latent^\top n\CovLike^{-1}\avgobs\right]\right)}\notag\\
    &=\exp\left(-\frac{1}{2}\left[\latent^\top\left(\CovApprox^{-1}-n\CovLike^{-1}\right)\latent-2\latent^\top\left(\CovApprox^{-1}\MeanPost-n\CovLike^{-1}\avgobs\right)\right]\right)\,.
\end{align}
If $\CovApprox^{-1}-n\CovLike^{-1}$ is positive definite, we can continue the computation
\begin{align*}
    \ypriorex(\latent)\propto_\latent \exp\left(-\frac{1}{2}\left[\latent^\top\left(\CovApprox^{-1}-n\CovLike^{-1}\right)\latent-2\latent^\top\left(\CovApprox^{-1}\MeanPost-n\CovLike^{-1}\avgobs\right)\right]\right)
    &\propto_\latent \exp\left(-\frac{1}{2}\left[\latent^\top\Covy^{-1}\latent-2\latent^\top\Covy^{-1}\Meany\right]\right)\\
    &\propto_\latent\Gaussarg{\latent}{\Meany}{\Covy}\,.
\end{align*}
If $\CovApprox^{-1}-n\CovLike^{-1}\eqqcolon S$ is not positive definite, then we can show that $\ypriorex$ is improper. In this case, $S$ has an eigenvalue $\lambda\leq 0$ and corresponding eigenvector $v\in\R^{d}$ with $\lnorm{v}=1$. Consider the hypercylinder $A$ of points whose distance to the axis $\R v$ is at most 1, formally defined as
\begin{align*}
    A\coloneqq\{\theta\in\R^d~|~\theta = tv+w,\text{ where }t\in\R, w\in v^\perp, \lnorm{w}=1\}\,,
\end{align*}
where $v^\perp=\{w\in\R^d~:~\scalprod{v}{w}=0\}$. Abbreviate $\gamma\coloneqq \CovApprox^{-1}\MeanPost-n\CovLike^{-1}\avgobs$ and collect all constants in $C>0$ (can change at different steps). Then we can lower bound the right hand side of Eq.~\eqref{eq:is_yprior_proper} on $A$ via
\begin{align*}
    \frac{\VIapprox(\latent|\exobs)}{\like(\exobs|\latent)}
    &= C \exp\left(-\frac{1}{2}\latent^\top S \latent+\scalprod{\theta}{\gamma}\right)\\
    &= C \exp\left(-\frac{1}{2}(tv+w)^\top S (tv+w)+\scalprod{tv+w}{\gamma}\right) \tag{$\theta=tv+w\in A$}\\
    &= C \exp\bigg(\underbrace{-\frac{1}{2}\lambda t^2}_{\geq 0} + \scalprod{v}{\gamma} t + \scalprod{w}{\gamma}-\frac{1}{2}w^\top S w\bigg) \tag{$Sv=\lambda v$, $\lnorm{v}=1$, $\scalprod{v}{w}=0$}\\
    &\geq C \exp\left(\scalprod{v}{\gamma} t + \scalprod{w}{\gamma}-\frac{1}{2}w^\top S w\right)\\
    &\geq C \exp\left(\scalprod{v}{\gamma} t \right)\tag{$\scalprod{w}{\gamma}-\frac{1}{2}w^\top S w$ is bounded for $\lnorm{w}\leq 1$}\,.
\end{align*}
Using this lower bound, we can lower bound the integral over $A$ through
\begin{align*}
    \int_A \frac{\VIapprox(\latent|\exobs)}{\like(\exobs|\latent)}\diff\latent
    \geq C\int_\R\exp\left(\scalprod{v}{\gamma} t \right)\diff t = \infty\,.
\end{align*}
Hence $\ypriorex$ is improper.

The last statement is that $\ypriorex$ is always proper in the setting \setprior\ ($\CovLike=I$), which means we need to show that $S$ has strictly positive eigenvalues.

We treat the cases \setreverse\ and \setforward\ separately. For \setreverse, it is
\begin{align*}
    S = \CovApprox^{-1}-n\CovLike^{-1}
      = \diag\left(\CovPrior^{-1}+nI\right) - nI
      =\diag\left(\CovPrior^{-1}\right)\,. \tag{$\diag$ is a linear operator}
\end{align*}
The diagonal entries of the symmetric positive definite matrix $\CovPrior^{-1}$ are lower bounded by its smallest eigenvalue $\lambda_\text{min}\left(\CovPrior^{-1}\right)>0$, which follows from the Courant–Fischer–Weyl min-max principle. Since these diagonal entries are the eigenvalues of $S$, this implies that $S$ is positive definite.
For the other case \setforward, we have
\begin{align*}
    S = \CovApprox^{-1}-n\CovLike^{-1}
    = \diag\left(\left(\CovPrior^{-1} + n\right)^{-1}\right)^{-1} - n\,.
\end{align*}
We again need to bound the diagonal elements of $\left(\CovPrior^{-1} + n\right)^{-1}$ with its eigenvalues. A similar argument as above yields
\begin{align}\label{eq:diag_upperbound_eigval}
    \lambda_\text{max}\left(\diag\left(\left(\CovPrior^{-1} + n\right)^{-1}\right)\right)
    \leq \lambda_\text{max}\left(\left(\CovPrior^{-1} + n\right)^{-1}\right)
    = \frac{1}{\lambda_\text{min}\left(\CovPrior^{-1}\right) + n}\,.
\end{align}
With this, the eigenvalues of $S$ are bouded by
\begin{align*}
    \lambda_\text{min}(S)
    = \lambda_\text{min}\left(\diag\left(\left(\CovPrior^{-1} + n\right)^{-1}\right)^{-1} - n\right)
    &= \frac{1}{\lambda_\text{max}\left(\diag\left(\left(\CovPrior^{-1} + n\right)^{-1}\right)\right)} - n\\
    &\geq \frac{1}{\frac{1}{\lambda_\text{min}\left(\CovPrior^{-1}\right) + n}} - n\tag{Eq.~\eqref{eq:diag_upperbound_eigval}}\\
    &= \lambda_\text{min}\left(\CovPrior^{-1}\right) > 0\,.
\end{align*}

\end{proof}

To prove Theorem~\ref{thm:gibbsmarg_ex} we require some general properties of Gaussian densities in Lemma~\ref{lem:Gaussian_props} and Lemma~\ref{lem:int_two_gaussians} because the proofs consist mainly of rearranging Gaussian densities. Next we compute the transition function of the Markov chain from Definition~\ref{def:gibbs_prior} in Proposition~\ref{prop:gibbs_transition}. We then prove Theorem~\ref{thm:gibbsmarg_ex} by guessing that the stationary distribution is Gaussian and verifying the stationary equation.

\begin{lemma}[Some properties of Gaussians]\label{lem:Gaussian_props}
    Let $\Gaussarg{x}{\mu}{\Sigma}$ denote the density of a Gaussian distribution $\mathcal{N}(\mu, \Sigma)$ on $\R^d$ at $x\in\R^d$ with mean $\mu\in\R^d$ and positive definite covariance $\Sigma\in\R^{d\times d}$. Then the following equalities hold
    \begin{enumerate}[label=(\roman*)]
        \item\label{item:mean_shift} $\Gaussarg{x+y}{\mu}{\Sigma}=\Gaussarg{x}{\mu-y}{\Sigma}$ and $\Gaussarg{x}{\mu}{\Sigma}=\Gaussarg{\mu}{x}{\Sigma}$ for $x,y\in\R^d$.
        \item\label{item:linear_trafo} Let $A\in\R^{d\times d}$ be non-singular. Then $\Gaussarg{Ax}{\mu}{\Sigma}=C_{A,\Sigma}\Gaussarg{x}{A^{-1}\mu}{A^{-1}\Sigma A^{-T}}$, where $C_{A,\Sigma}\in\R$ is a constant that depends only on $A$ and $\Sigma$.
        \item\label{item:convolution} Let $\mu_1,\mu_2\in\R^d$ and $\Sigma_1,\Sigma_2\in\R^{d\times d}$ positive definite. Then the convolution of two Gaussian densities corresponds to the sum of two independent Gaussians, i.e., for $z\in\R^d$ it holds
        \begin{align*}
            \int_{\R^d}\Gaussarg{z-x}{\mu_1}{\Sigma_1}\Gaussarg{x}{\mu_2}{\Sigma_2}\diff x
            = \left[\Gaussarg{\cdot}{\mu_1}{\Sigma_1}\ast \Gaussarg{\cdot}{\mu_2}{\Sigma_2}\right](z)
            =\Gaussarg{z}{\mu_1+\mu_2}{\Sigma_1+\Sigma_2}\,.
        \end{align*}
    \end{enumerate}
\end{lemma}
\begin{proof}
Points $\ref*{item:mean_shift}$ and $\ref*{item:convolution}$ are trivial. For $\ref*{item:linear_trafo}$, we compute
\begin{align*}
    \Gaussarg{Ax}{\mu}{\Sigma}
    = C_\Sigma\exp\left(-\frac{1}{2}(Ax-\mu)^\top \Sigma^{-1}(Ax-\mu)\right)
    &= C_\Sigma\exp\left(-\frac{1}{2}(x-A^{-1}\mu)^\top A^\top \Sigma^{-1}A(x-A^{-1}\mu)\right)\\
    &= C_{A,\Sigma}\Gaussarg{x}{A^{-1}\mu}{A^{-1}\Sigma A^{-T}}\,.
\end{align*}
Note that $A^{-1}\Sigma A^{-T}$ is positive definite as well: symmetry is obvious and it holds
\begin{align*}
    x^\top A^{-1}\Sigma A^{-T} x =(A^{-T}x)^\top \Sigma(A^{-T}x)> 0
\end{align*}
for $x\neq 0$ because $A$ is non-singular and $\Sigma$ is positive definite.
\end{proof}

The next lemma computes an integral that appears both in computing the transition function of the Gibbs chain and in computing its stationary distribution.
\begin{lemma}\label{lem:int_two_gaussians}
    Let $\latent^\prime, a,\mu_2\in\R^d$, $A\in\R^{d\times d}$ non-singular, and $\Sigma_1,\Sigma_2\in\R^{d\times d}$ positive definite. Then it is
    \begin{align*}
        \int_{\R^d}\Gaussarg{\latent^\prime}{a+Ax}{\Sigma_1}\Gaussarg{x}{\mu_2}{\Sigma_2}\diff x
        = \Gaussarg{\latent^\prime}{a+A\mu_2}{\Sigma_1+A\Sigma_2 A^\top }\,.
    \end{align*}
\end{lemma}
\begin{proof}
We start by rephrasing the first density
\begin{align*}
    \Gaussarg{\latent^\prime}{a+Ax}{\Sigma_1}
    &=\Gaussarg{Ax}{\latent^\prime-a}{\Sigma_1} \tag{Lemma~\ref{lem:Gaussian_props}, $\ref*{item:mean_shift}$}\\
    &=C_{A,\Sigma_1}\Gaussarg{x}{A^{-1}(\latent^\prime-a)}{A^{-1}\Sigma_1 A^{-T}}\tag{Lemma~\ref{lem:Gaussian_props}, $\ref*{item:linear_trafo}$}\\
    &=C_{A,\Sigma_1}\Gaussarg{A^{-1}\latent^\prime - x}{A^{-1}a}{A^{-1}\Sigma_1 A^{-T}}\tag{Lemma~\ref{lem:Gaussian_props}, $\ref*{item:mean_shift}$}\,.
\end{align*}
This yields
\begin{align*}
     \int_{\R^d}\Gaussarg{\latent^\prime}{a+Ax}{\Sigma_1}\Gaussarg{x}{\mu_2}{\Sigma_2}\diff x
     &\propto_\latent^\prime \int_{\R^d}\Gaussarg{A^{-1}\latent^\prime - x}{A^{-1}a}{A^{-1}\Sigma_1 A^{-T}}\Gaussarg{x}{\mu_2}{\Sigma_2}\diff x \\
     &=\Gaussarg{A^{-1}\latent^\prime}{A^{-1}a+\mu_2}{A^{-1}\Sigma_1A^{-T}+\Sigma_2}\tag{Lemma~\ref{lem:Gaussian_props}, $\ref*{item:convolution}$}\\
     &\propto_\latent^\prime\Gaussarg{\latent^\prime}{a+A\mu_2}{\Sigma_1+A\Sigma_2A^\top }\tag{Lemma~\ref{lem:Gaussian_props}, $\ref*{item:linear_trafo}$}\,.
\end{align*}
\end{proof}

We can now compute the transition function of the Gibbs chain from Definition~\ref{def:gibbs_prior}.
\begin{proposition}\label{prop:gibbs_transition}
The transition function of the Gibbs chain is given by Gaussian distributions
\begin{align}\label{eq:gibbs_transition}
    \transit(\latent^\prime|\latent) = \Gaussarg{\latent^\prime}{a+A\latent}{B}\,,
\end{align}
where $\latent,\latent^\prime\in\R^d$, $a=\CovPost\CovPrior^{-1}\MeanPrior$, $A=n\CovPost\CovLike^{-1}$, and $B=\CovApprox+n\CovPost\CovLike^{-1}\CovPost$.
\end{proposition}
\begin{proof}
Let $\latent,\latent^\prime\in\R^d$. By definition, the transition function of the Gibbs chain is given by
\begin{align*}
    r(\latent^\prime|\latent)=\int_{\R^{n\times d}}\VIapprox(\latent^\prime|\exobs)\like(\exobs|\latent)\diff \exobs
    =\int_{\R^d}q(\latent^\prime|\avgobs)\avglike(\avgobs|\latent)\diff\avgobs\,,
\end{align*}
where we have transformed the integral to the mean $\avgobs$, because $\VIapprox$ only depends on $\exobs$ through $\avgobs$. The corresponding push-forward measure is given by $\avglike(\avgobs|\latent)=\Gaussarg{\avgobs}{\latent}{1/n\CovLike}$. Using Proposition~\ref{prop:distributions_ex} and expressing $\MeanPost$ with $a$ and $A$, we have that
$q(\latent^\prime|\avgobs)=\Gaussarg{\latent^\prime}{\MeanPost}{\CovApprox}=\Gaussarg{\latent^\prime}{a+A\avgobs}{\CovApprox}$. Putting everything together, we get
\begin{align*}
    r(\latent^\prime|\latent) &= \int_{\R^d}\Gaussarg{\latent^\prime}{a+A\avgobs}{\CovApprox}\Gaussarg{\avgobs}{\latent}{\frac{1}{n}\CovLike}\diff\avgobs\\
    &= \Gaussarg{\latent^\prime}{a+A\latent}{\CovApprox + A\frac{1}{n}\CovLike A^\top }\tag{Lemma~\ref{lem:int_two_gaussians}}\,.
\end{align*}
The equality $A\frac{1}{n}\CovLike A^\top=n\CovPost\CovLike^{-1}\CovPost$ concludes the proof.
\end{proof}

We are now ready to prove Theorem~\ref{thm:gibbsmarg_ex}.

\begin{proof}[Proof of Theorem~\ref{thm:gibbsmarg_ex}]
The proof is based on guessing that the stationary distribution is Gaussian. We first show that Gaussian distributions are closed under taking a step with the transition function and then derive the parameters of the stationary distribution based on the stationary equation.

Let $p(\latent)=\Gaussarg{\latent}{m}{M}$ with $m\in\R^d$ and $M\in\R^{d\times d}$ positive definite. Using Lemma~\ref{lem:int_two_gaussians} and Proposition~\ref{prop:gibbs_transition}, the distribution after one step $Rp$ is given by
\begin{align}\label{eq:trans_dist_ex}
    Rp(\latent^\prime)
    =\int_{\R^d}\transit(\latent^\prime|\latent)p(\latent)\diff\latent
    =\int_{\R^d}\Gaussarg{\latent^\prime}{a+A\latent}{B}\Gaussarg{\latent}{m}{M}\diff\latent
    =\Gaussarg{\latent^\prime}{a+Am}{B+AMA^\top }\,.
\end{align}
If a $p$ satisfies the stationary equation $p=Rp$, then it is the stationary distribution $p=\Gibbsprior$. Using that $Rp$ is again Gaussian, Eq.~\eqref{eq:trans_dist_ex} shows that this is satisfied if and only if
\begin{align*}
    m=a+Am
    \qand
    M = B+AMA^\top \,.
\end{align*}
The solution for the mean equation is obtained by rearranging and plugging in the definitions of $a, A$ and $\CovPost$:
\begin{align*}
    m
    =\left(I-A\right)^{-1}a
    =\left(I-n\CovPost\CovLike^{-1}\right)^{-1}\CovPost\CovPrior^{-1}\MeanPrior
    &=\left(\CovPost^{-1}-n\CovLike^{-1}\right)\CovPrior^{-1}\MeanPrior\\
    &=\left(\CovPrior^{-1}+n\CovLike^{-1}-n\CovLike^{-1}\right)^{-1}\CovPrior^{-1}\MeanPrior\\
    &=\MeanPrior\,.
\end{align*}
The stationary equation for the covariance matrix $M$ is equivalent to the Lyapunov equation
\begin{align*}
    AM A^\top - M + B = 0\,,
\end{align*}
which has a unique solution. This concludes the proof.
\end{proof}

\subsection{Numerical Evaluation of the Biases in the Gaussian Toy Example}\label{app:toy_bias_numerical}
This section presents numerical values that back up the statements about the biases from Section~\ref{sec:toy_bias}.

The first bias is the compactness of the mean-field approximations. Table~\ref{tab:bias_compactness} shows the compactness of all relevant distributions as measured by the entropy, which is given by $d/2(1+\ln(2\pi))+1/2\ln(\det\Sigma)$ for a Gaussian distribution $\Gauss(\mu,\Sigma)$.
As stated in the main paper, under the setting \setforward\ approximations $\VIapprox$ are less compact than the exact posterior $\posterior$. This is reflected by the priors as the Gibbs prior $\Gibbsprior$ is less compact than the exact prior $\prior$.
Under the setting \setreverse, this trend is reversed: approximations are more compact than the exact posterior, and the Gibbs prior is more compact than the exact prior.

The second bias is the loss of correlation under the mean-field approximations. Table~\ref{tab:bias_correlation} shows the correlation between different components $\latent_1$ and $\latent_2$ of the 2-dimensional latent variable $\latent=(\latent_1,\latent_2)$ only for the prior distributions, because the components of the approximations are by definition uncorrelated.
Recall that the exact posterior distribution was the same in both settings, but under \setprior\ the posterior correlation was due to prior correlation, whereas under \setlike\ it was due to likelihood correlation.
In the setting \setprior, the Gibbs prior is less correlated than the prior. In the setting \setlike, the prior is uncorrelated, but the Gibbs prior is negatively correlated to ``cancel out'' the positive correlation of the likelihood covariance.

\setlength{\tabcolsep}{10pt}
\begin{table}[ht]
\caption{Compactness of various distributions across settings as measured by the entropy. First value is under the setting \setprior\ and second value is under the setting \setlike. Note that the covariance of exact and approximate posterior does not depend on the observation.}
\label{tab:bias_compactness}
\vskip 0.15in
\begin{center}
\begin{small}
\begin{tabular}{lcc}
\toprule
Entropy& \setforward & \setreverse\\
\midrule
Prior $\prior$ & 2.24 / 2.84 & 2.24 / 2.84\\
Gibbs prior $\Gibbsprior$ & 2.82 / 3.15 & 2.21 / 2.52\\
Exact posterior $\posterior$ & 1.50 / 1.50 & 1.50 / 1.50\\
Approximate posterior $\VIapprox$ & 1.97 / 1.97 & 1.02 / 1.02\\
\bottomrule
\end{tabular}
\end{small}
\end{center}
\vskip -0.1in
\end{table}

\setlength{\tabcolsep}{10pt}
\begin{table}[ht]
\caption{Correlation of prior distributions across settings as measured by the covariance $\Cov(\latent_1,\latent_2)$ between the components of $\latent=(\latent_1,\latent_2)\in\R^2$. First value is under the prior distribution $\prior$ and second value is under the corresponding Gibbs prior $\Gibbsprior$.}
\label{tab:bias_correlation}
\vskip 0.15in
\begin{center}
\begin{small}
\begin{tabular}{lcc}
\toprule
Covariance $\Cov(\latent_1,\latent_2)$& \setforward & \setreverse\\
\midrule
\setprior & 1.45 / 0.91 & 1.45 / 0.74\\
\setlike & 0 / -1.13 & 0 / -0.52\\
\bottomrule
\end{tabular}
\end{small}
\end{center}
\vskip -0.1in
\end{table}

\section{EXPERIMENTAL DETAILS}\label{app:exp}
In this section, we give more details on the Bayesian models and approximations to their posteriors which are considered in Section~\ref{sec:experiments}. We used the python library numpyro \citep{Pha:2019} for the posterior approximation methods Laplace, NUTS, and ADVI.

\paragraph{Baseline \citet{Tal:2018}} We allocate this baseline the same resources as the corresponding Gibbs chain in terms of draws from the posterior. That is, if our Gibbs chain runs for $M$ steps, the baseline repeats $N$ draws $(\tilde{\latent},\tilde{\obs})$ with $\latent_1,\dotsc,\latent_L\sim\VIapprox(\cdot|\tilde{\obs})$ such that $N\cdot L\approx M$. 
Specifically, we choose $N=323$ and $L=31$. For the histograms in Figure~\ref{fig:baseline}, we re-binned once to reduce noise.

\paragraph{Convergence monitoring}
\begin{figure*}[t]
    \centering
    \includegraphics{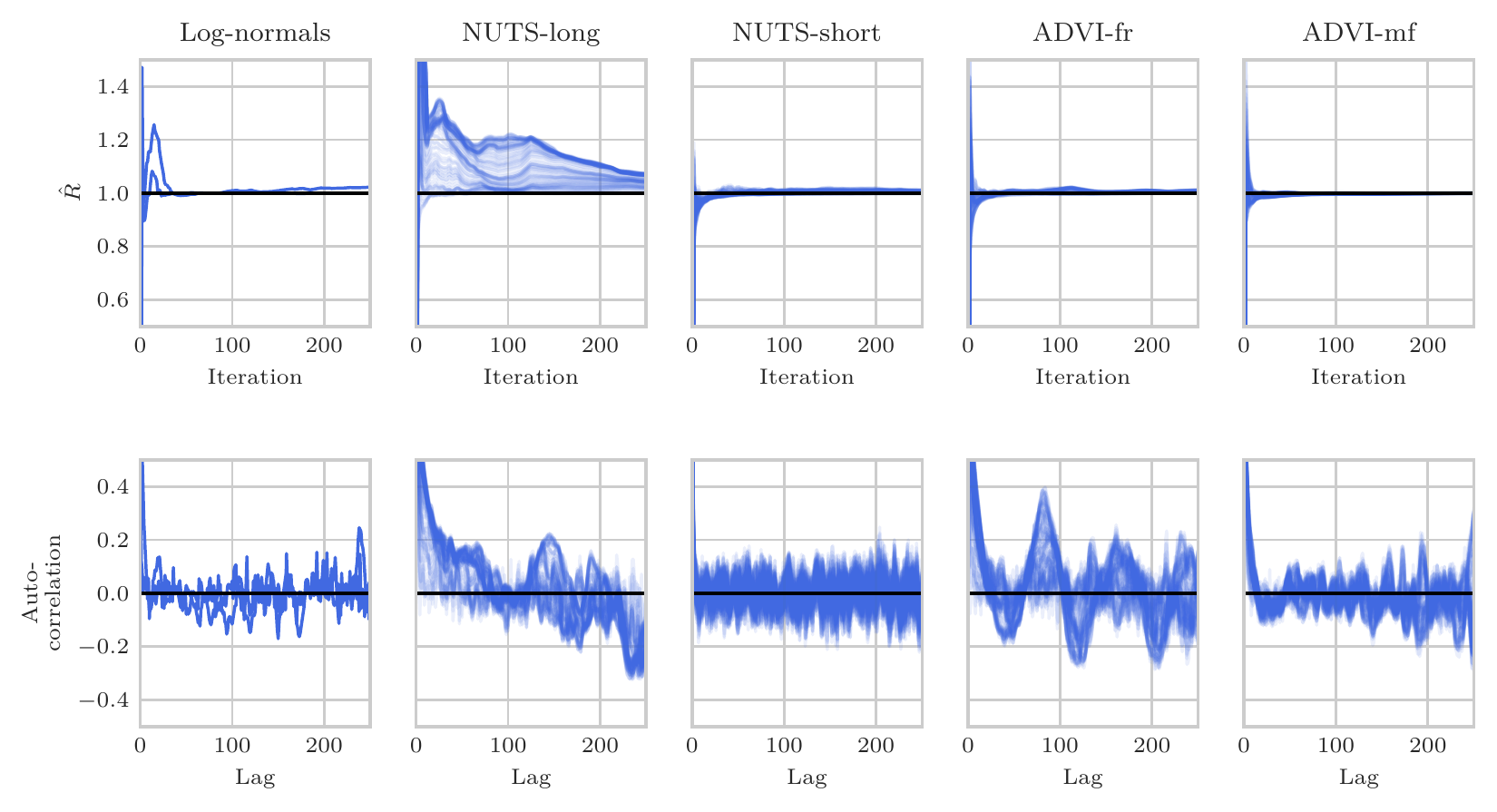}
    \caption{Gelman-Rubin diagnostic $\hat{R}$ (\textbf{top row}) and lag-$k$ autocorrelation (\textbf{bottom row}) for the Gibbs chains from Section~\ref{sec:experiments} with one curve per dimension. Section~\ref{sec:exp_log_normals} (\textbf{first column}) has $d=2$ dimensions and Section~\ref{sec:exp_volatility} (\textbf{other columns}) have $d=100$ dimensions. Values $\hat{R}\approx 1$ or lag-$k$ autocorrelation $\approx 0$ indicate convergence of the Gibbs chain.
    }
    \label{fig:monitoring}
\end{figure*}

We monitor the convergence of our Gibbs chains with two standard measures, the Gelman-Rubin diagnostic $\hat{R}$ \citep{Gel:1992} and the lag-$k$ autocorrelation. Both are shown in Figure~\ref{fig:monitoring} for all experiments from Section~\ref{sec:experiments}.
The Gelman-Rubin diagnostic $\hat{R}$ uses multiple chains to compute the ratio of between-chain variance to within-chain variance. A ratio $\hat{R}\approx 1$ indicates convergence. The top row of Figure~\ref{fig:monitoring} shows that this value is reached quickly in all cases except for NUTS-long, which takes longer to converge. Potential explanations are that convergence is generally slower in the high-dimensional setting ($d=100$ for NUTS-long compared to $d=2$ for Log-normals) and that the other less accurate methods 
introduce additional bias that promotes faster convergence.
The lag-$k$ autocorrelation is defined as the correlation of a sequence with its shifted version by $k$ steps. A high autocorrelation of a Markov chain indicates slow mixing and thus slower convergence. The bottom row of Figure~\ref{fig:monitoring} shows the autocorrelation for the Gibbs chains, which eventually oscillate around 0 due to finite sample noise.
The autocorrelation gets close to 0 quickly for the low-dimensional setting log-normals and for the less accurate methods NUTS-short and ADVI-mf in the high-dimensional setting. Only the more accurate methods in the high-dimensional setting NUTS-long and ADVI-fr take longer to reach 0. This hints towards slower convergence and is in line with \citet{Tal:2018}, who predict slow convergence of the Gibbs chain when the parameters are strongly correlated to the observations. In particular, we expect this to be the case for a large number of observations. 

\subsection{Sum of log-normals}
The Bayesian model has latent parameters $\latent=(\mu,\sigma^2)\in\R\times\R_{>0}$, on which we place a prior $\prior(\latent)$ with independent marginal distributions $\mu\sim\Gauss(0, 1)$ and $\sigma^2\sim\text{Gamma}(1, 1)$. The likelihood $\like(\obs|\latent)$ for an observation $\obs>0$ is given by an $L$-fold convolution of a log-normal distribution, that is, $\obs|\latent\sim\LogNormal^{\ast L}(\mu,\sigma^2)$.

To obtain the approximation $\VIapprox(\latent|\obs)$ to the true posterior $\posterior(\latent|\obs)$ of this model, we employ the following two-step procedure:
\begin{enumerate}
    \item Define an approximate likelihood $\tilde{\like}(\obs|\latent)$ as the Fenton-Wilkinson approximation to the true likelihood $\like$, which is another log-normal distribution with matching first two moments. Specifically, $\tilde{\like}(\cdot|\latent)$ describes the distribution $\LogNormal(\alpha,\beta^2)$, where
    \begin{align*}
        \alpha &= \mu + \log L + 0.5\left(\sigma^2 - \beta^2\right)\,,\\
        \beta^2 &= \log\left[\frac{\exp{\sigma^2}-1}{L} + 1\right]\,.
    \end{align*}
    \item Define $\VIapprox(\latent|\obs)$ as the Laplace-approximation to the posterior of this new model $\tilde{\posterior}(\latent|\obs)\propto_\latent \prior(\latent)\tilde{\like}(\obs|\latent)$. This means that $\VIapprox(\cdot|\obs)$ describes a bivariate normal distribution $\Gauss(\latent_\obs^\ast, \Sigma_\obs)$ with
    \begin{align*}
        \latent_\obs^\ast &= \argmax_\latent \prior(\latent)\tilde{\like}(\obs|\latent)\\
        \Sigma_\obs &= -H_{\log\tilde{\posterior}}^{-1}\,,
    \end{align*}
    where $H_{\log\tilde{\posterior}}$ describes the Hessian matrix of $\latent\mapsto\log\left(\prior(\latent)\tilde{\like}(\obs|\latent)\right)$.
\end{enumerate}

\subsection{Stochastic Volatility}
This model is a simplified model of the one described in \citet{Hof:2014}, who place additional prior distributions on the parameters $\sigma$, $\nu$, and $\theta_0$. We made the simplifying choice $\theta_0=0$. The other hyperparameters $\sigma=.09$ and $\nu=12$ were chosen by taking the posterior means under S\&P500 dataset. The posterior was approximated with NUTS where the priors were \(\sigma \sim \mathrm{Exp}(50)\) and \(\nu \sim \mathrm{Exp}(0.1)\), following \citet{Hof:2014}.

\paragraph{Measuring compactness and divergence}
Table~\ref{tab:volatility} supplements our statements about the bias of the approximation methods in Section~\ref{sec:exp_volatility}.
Regarding compactness, we can confirm that the methods NUTS-short, ADVI-fr, and ADVI-mf  are overly compact compared to the original prior.
Regarding divergence, we see that the method NUTS-long is closest to the original prior and the restrictive method ADVI-mf is farthest. The more powerful versions yield Gibbs priors that are closer to the original prior, that is, NUTS-long is closer than NUTS-short and ADVI-fr is closer than ADVI-mf.
\setlength{\tabcolsep}{10pt}
\begin{table}[ht]
\caption{Compactness and distance to original prior for the approximation methods of Section~\ref{sec:exp_volatility}. Compactness is measured by the Frobenius norm of the empirical covariance matrix and distance to the original prior is measured by the maximum mean discrepancy under the Gaussian kernel $k(x,y)=\exp(-\lnorm{x-y}^2/2)$.}
\label{tab:volatility}
\vskip 0.15in
\begin{center}
\begin{small}
\begin{tabular}{lccccc}
\toprule
& Original prior & NUTS-long & NUTS-short & ADVI-fr & ADVI-mf\\
\midrule
Compactness & 34.18 & 23.81 & 3.37 & 6.09 & 1.21 \\
Distance to original prior & 0 & 0.014 & 0.035 & 0.021 & 0.179 \\
\bottomrule
\end{tabular}
\end{small}
\end{center}
\vskip -0.1in
\end{table}

\end{document}